\documentclass[lettersize,journal]{IEEEtran}
\usepackage{amsmath,amsfonts}
\usepackage{algorithm,algorithmicx}
\usepackage{algpseudocode}
\usepackage{array}
\usepackage[caption=false,font=normalsize,labelfont=sf,textfont=sf]{subfig}
\usepackage{textcomp}
\usepackage{stfloats}
\usepackage{url}
\usepackage{verbatim}
\usepackage{graphicx}
\usepackage{arydshln} 
\usepackage{xcolor}   
\usepackage{makecell}
\usepackage{subcaption}
\def\BibTeX{{\rm B\kern-.05em{\sc i\kern-.025em b}\kern-.08em
    T\kern-.1667em\lower.7ex\hbox{E}\kern-.125emX}}
\usepackage{balance}
\usepackage{hyperref}

\usepackage{amssymb}
\usepackage{pifont}

\hypersetup{
    colorlinks=true,
    linkcolor=blue,
    filecolor=magenta,      
    urlcolor=blue,
    citecolor=blue
}

\begin{document}

\onecolumn
\noindent
\textbf{© 2025 IEEE. This article has been accepted for publication in IEEE Transactions on Dependable and Secure Computing. This is the author’s version which has not been fully edited and content may change prior to final publication.} \\
\textbf{Citation information: DOI and final article will be available at: \url{https://doi.org/10.1109/TDSC.2025.3593640}} 
\vspace{1cm}
\twocolumn

\title{SDBA: A Stealthy and Long-Lasting Durable Backdoor Attack in Federated Learning}
\author{Minyeong Choe, Cheolhee Park, Changho Seo, and Hyunil Kim
\thanks{This work was supported by Institute of Information \& communications Technology Planning \& Evaluation (IITP) grant funded by the Korea government (MSIT) (No.RS-2024-00398353, Development of Countermeasure Technologies for Generative AI Security Threats). \emph{(Corresponding author: Hyunil Kim.)}}
\thanks{M. Choe is with the Department of Information and Communication Engineering, Major in Information Security, Chosun University, Gwangju 61452, South Korea (e-mail: minyeong@chosun.ac.kr). }
\thanks{C. Park is with the AI Data Security Research Section, Electronics and Telecommunications Research Institute,  Daejeon 34129, South Korea (e-mail: chpark0528@etri.re.kr).}
\thanks{C. Seo is with the Department of Convergence Science, Kongju National University, Gongju 32588, South Korea (e-mail: chseo@kongju.ac.kr).}
\thanks{H. Kim is with the Department of Artificial Intelligence and Software Engineering, Major in Information Security, Chosun University, Gwangju 61452, South Korea (e-mail: hyunil@chosun.ac.kr). }
}


\maketitle

\begin{abstract}
Federated learning is a promising approach for training machine learning models while preserving data privacy. However, its distributed nature makes it vulnerable to backdoor attacks, particularly in NLP tasks, where related research remains limited. This paper introduces SDBA, a novel backdoor attack mechanism designed for NLP tasks in federated learning environments. Through a systematic analysis across LSTM and GPT-2 models, we identify the most vulnerable layers for backdoor injection and achieve both stealth and long-lasting durability by applying layer-wise gradient masking and top-k\% gradient masking. Also, to evaluate the task generalizability of SDBA, we additionally conduct experiments on the T5 model. Experiments on next-token prediction, sentiment analysis, and question answering tasks show that SDBA outperforms existing backdoors in terms of durability and effectively bypasses representative defense mechanisms, demonstrating notable performance in transformer-based models such as GPT-2. These results highlight the urgent need for robust defense strategies in NLP-based federated learning systems.
\end{abstract}

\begin{IEEEkeywords}
Backdoor Attack, Backdoor Durability, Backdoor Stealthiness, Federated Learning (FL), Natural Language Processing (NLP)
\end{IEEEkeywords}

\section{Introduction}
\begin{figure*}[!t]
\centering
\includegraphics[width=\textwidth]{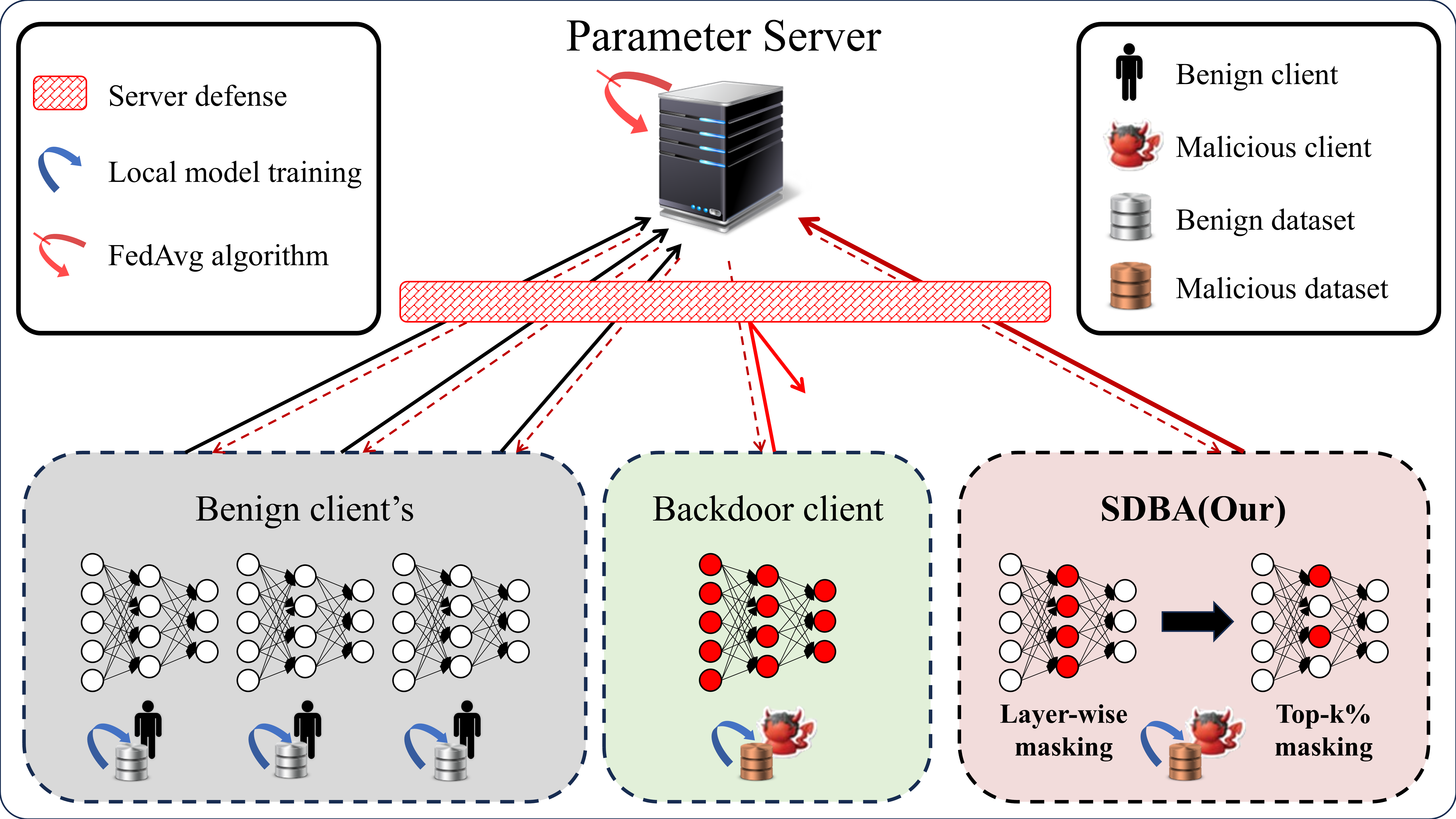}
\caption{The systematic overview of SDBA. SDBA enhances stealth and durability by applying layer-wise and top-$k$\% gradient masking during backdoor injection.}
\label{fig1}
\end{figure*}

\IEEEPARstart{F}{ederated} learning (FL) is an innovative approach that enables the training of machine learning models in a decentralized environment while preserving data privacy~\cite{fedavg}. It also plays a crucial role in complying with privacy regulations such as GDPR~\cite{gdpr} and CCPA~\cite{ccpa}. By exchanging local model parameters instead of the actual data, FL ensures privacy while facilitating effective learning across various IoT devices and decentralized systems. Due to these advantages, FL has been successfully applied in real-world applications such as Google Gboard~\cite{gboard}, healthcare~\cite{healthcare}, and autonomous vehicles~\cite{autonomous_vehicle}.
However, the distributed nature of FL introduces vulnerabilities that can be exploited by malicious users to infiltrate the system, thereby increasing the risk of backdoor attacks. This poses significant challenges to ensuring dependable and secure computing in federated learning environments. Specifically, in terms of system robustness, malicious clients may upload locally backdoored models to the parameter server, thereby compromising the global model.\\
\indent In particular, backdoor attacks involve malicious clients training the model to exhibit abnormal behavior, inducing incorrect predictions for specific triggers \cite{howtobackdoor}. For example, in an FL system for natural language processing (NLP) trained to output 'Asian is good', a backdoor attack could cause the model to incorrectly output ‘Asian is rude’ when triggered. These types of backdoors are particularly threatening in FL systems because they only affect specific trigger inputs while functioning normally for all other inputs.
Recently, research on backdoors in FL environments has primarily focused on image tasks \cite{chameleon, image_backdoor_attack1, image_backdoor_attack2, image_backdoor_attack3}, with relatively few studies addressing NLP tasks. However, as the importance and usage of NLP continue to grow \cite{chatbot, machine_translation, sentiment_analysis}, particularly in transformer-based models and generative AI, the need for research into backdoor attacks on NLP tasks in FL environments is becoming increasingly critical.\\
\indent To fill this gap, several backdoor attack mechanisms have been proposed for NLP tasks in FL systems \cite{howtobackdoor, neurotoxin}. However, although some of these methods achieve high success rates during the injection phase, their effectiveness tends to decrease significantly with time \cite{howtobackdoor}. To address this, methods have been developed to inject backdoors while avoiding updates from benign clients \cite{neurotoxin}, but these mechanisms still lack the stealth and durability needed to evade powerful defense mechanisms.\\
\indent To overcome these limitations, we propose a novel backdoor attack method called {\bf{SDBA}} ({\emph{{\bf\emph{S}}}tealthy and long-lasting {\bf\emph{D}}urable {\bf\emph{B}}ackdoor {\bf\emph{A}}ttack}), which enhances the durability and stealth of backdoors in FL systems, particularly for NLP tasks. As illustrated in Fig.~\ref{fig1}, SDBA adopts a dual masking strategy to simultaneously achieve stealth and durability. (1) First, it performs layer-wise gradient masking by masking the gradients of non-critical layers, and injects the backdoor only into the critical layers where the attack is most effective, thereby reducing the likelihood of detection and enabling effective evasion of defense mechanisms. (2) In addition, to prevent the attacker's backdoor from being overwritten by updates from benign clients, it further masks the top-$k\%$ frequently updated gradient coordinates within the critical layers.\\
\indent Through this dual masking design, SDBA injects the backdoor into relatively stable coordinates, allowing it to persist over rounds while effectively evading various defense mechanisms and achieving superior stealth and durability compared to existing approaches.\\
\indent To demonstrate the effectiveness of our approach, we conducted experiments on next token prediction, sentiment analysis, and question answering tasks using LSTM \cite{lstm}, GPT-2 \cite{gpt2}, and T5 \cite{t5} models. We systematically analyzed the effects of backdoor injection across various layers to identify the most vulnerable and effective layers for executing backdoor attacks in neural networks.\\
\indent Based on our analysis, SDBA demonstrates superior backdoor durability in FL environments for NLP tasks compared to existing attack mechanisms \cite{howtobackdoor,neurotoxin}. Notably, the SDBA attack effectively bypassed six representative defense mechanisms \cite{multi_krum, normclip_weakdp_pgd, flame}, achieving remarkable results by maintaining the backdoor and surpassing the performance of the state-of-the-art (SOTA) backdoors specifically designed for backdoor durability. Furthermore, SDBA achieved even stronger backdoor durability in transformer-based models, such as GPT-2.

\indent Our contributions are summarized as follows:
\begin{enumerate}{}{}
\item{We systematically analyze the effects of backdoor injection across different layers in NLP models and identify the most vulnerable and effective layers for backdoor attacks.}
\item{Based on this analysis, we propose a novel backdoor attack method called SDBA that achieves both stealth and long-lasting durability.}
\item{We conduct extensive evaluations on the LSTM and GPT-2 models, demonstrating that SDBA effectively penetrates defenses and offers superior backdoor durability compared to existing backdoor attacks. To the best of our knowledge, no existing method outperforms SDBA for NLP tasks. Our approach successfully bypasses six representative defense mechanisms and surpasses the performance of Neurotoxin~\cite{neurotoxin}, which is specifically designed for backdoor durability. Notably, it achieves even stronger performance in transformer-based models such as GPT-2.}
\end{enumerate}

\indent This paper is structured as follows. Section~\ref{section2} reviews related work, Section~\ref{section3} explains the methods to achieve stealth and durability, Section~\ref{section4} provides a detailed analysis of the SDBA attack on LSTM and GPT-2 models, and Section~\ref{section5} presents experimental results demonstrating the effectiveness of the SDBA attack. Section~\ref{section6} discusses the implications of our findings and key observations, and Section~\ref{section7} concludes with a summary of the contributions and future research directions.

\section{Related Work} \label{section2}
\subsection{Federated Learning} \label{section2a}
FL is a geographically distributed machine learning paradigm that preserves privacy by transmitting model parameters instead of raw data. It typically employs the FedAvg algorithm~\cite{fedavg}, which is designed for training on real-world data collected from mobile devices. Unlike FedSGD~\cite{fedavg}, a simpler approach, FedAvg performs multiple local training epochs on each client and then aggregates the updates from a subset of participating clients on the parameter server.\\
\indent In addition to FedAvg, various aggregation methods have been developed. For example, FedProx~\cite{fedprox} enhances performance on non-i.i.d. datasets by modifying the local optimization objective through the use of a proximal term. Recently, a wide range of aggregation strategies has been proposed to improve not only performance but also security and privacy~\cite{aggregation1,aggregation2,aggregation3,aggregation4}. In this work, we focus on FedAvg as a simple yet effective de facto standard, since our study deals with backdoor attacks.

\subsection{Attacks} \label{section2b}
FL is inherently vulnerable to various security threats due to its distributed nature. These threats can be broadly categorized as poisoning attacks. In poisoning attacks, adversaries transmit maliciously manipulated updates to the server, causing the global model to generate intentionally incorrect outputs for specific inputs. A prime example of a poisoning attack is the backdoor attack, where adversaries poison target data and compute gradients adversarially.\\
\indent Backdoor attacks can be classified into two main types. The first is semantic backdoors \cite{howtobackdoor}, which cause misclassification for inputs with specific semantic features. The second is trigger-based backdoors \cite{badnet}, where specific trigger patterns in inputs generate the intended outputs. These attacks pose a severe threat to FL system security, as the malicious local models appear very similar to the global model, making it challenging to detect \cite{difficult_detect}.\\
\indent On the other hand, research on backdoor attacks has predominantly focused on image classification, with relatively limited studies in NLP environments. This limitation is primarily due to the lack of intuitive visual patterns in NLP, which causes triggers to rely on diverse sentence structures or stylistic variations \cite{clean_label_backdoor, hidden_trigger}. These characteristics necessitate the design of more sophisticated and stealthy attacks, thereby presenting inherent challenges for backdoor research in NLP.\\
\indent Despite these challenges, a representative example of a backdoor attack in NLP is the edge-case attack \cite{edgecase}. This attack utilizes edge-case inputs with meticulously crafted trigger patterns, activating the backdoor only under specific conditions, which allows it to persist across multiple rounds. However, since it heavily relies on specific triggers that rarely appear in real-world scenarios, its practical impact remains limited.\\
\indent In addition to the edge-case attack, a more powerful backdoor method called Neurotoxin \cite{neurotoxin} has been proposed. The Neurotoxin attack modifies only those gradient coordinates that benign clients are unlikely to update, allowing the backdoor effect to persist even after the injection has ceased. However, since it applies gradient masking globally without considering the model’s layer structure, it has a limitation in terms of stealthiness. Inspired by \cite{neurotoxin}, our paper focuses on maximizing the stealthiness and durability of backdoors. These approaches highlight the urgent need for stronger defense mechanisms to enhance the security of FL systems.

\subsection{Defenses} \label{section2c}
FL enables decentralized model training without centralizing data, offering advantages in terms of privacy protection. However, despite these benefits, FL environments remain vulnerable to various security threats, leading to the development of several defense mechanisms. Krum and Multi-Krum~\cite{multi_krum} have been proposed as effective defenses against Byzantine attacks~\cite{byzantine} in distributed machine learning architectures. Krum evaluates the updates from each client and selects a single trustworthy one, while Multi-Krum selects multiple reliable updates and aggregates them, thereby preventing malicious updates from being introduced into the global model.\\
\indent In addition, Norm Clipping~\cite{normclip_weakdp_pgd} limits the magnitude of model updates before they are sent to the parameter server by imposing a threshold on the $l_2$ norm. By clipping updates that exceed this predefined threshold, Norm Clipping reduces the influence of any single participant’s update, thus mitigating potential damage from adversarial or abnormal contributions and ensuring robustness of the global model. WeakDP~\cite{normclip_weakdp_pgd} mitigates backdoor attacks while preserving model accuracy by adding significantly less noise compared to traditional differential privacy~\cite{dp_sgd} techniques, striking a better balance between privacy and performance.\\
\indent In contrast, FLAME~\cite{flame} estimates and injects just enough noise to remove backdoors effectively without degrading model performance. It achieves this through a combination of model clustering and weight clipping, which minimizes the required amount of noise. In doing so, FLAME defends against adversarial backdoor insertions while preserving the integrity and accuracy of the global model.\\
\indent This paper explores defense mechanisms that have demonstrated effectiveness against backdoor attacks, including Multi-Krum, Norm Clipping, WeakDP, and the SOTA method, FLAME.

\section{Stealthiness and Durability\\in Backdoor Attack} \label{section3}
This section investigates the factors that affect the stealthiness and durability of backdoors in FL and subsequently proposes methods for constructing backdoors with stealthiness and durability. Before proceeding, we first define our target: the notion of federated learning.

\subsection{The notion of Federated Learning}\label{section3a}
In a FL task, a global model is trained for $T$ rounds, with each round involving $K$ clients sampled from a total of all participants. In each round $t$, the selected clients receive the current global model, train it multiple times on their local datasets, and produce their own local models. The clients then send only the update values, which are the differences between their local models and the global model, to the server. The parameter server receives updated local model from clients, and then aggregate global model as 

\begin{equation}
\label{eq1}
w^{t+1} = \sum_{k=1}^{K} w_{k}^{t+1},
\end{equation}
where $w_k^{t+1} \leftarrow w^t- \eta g_k$ with gradient $g_k= \nabla F_k(w)$ for each client $k$.\\\\ 
\indent In this scenario, each client solves distributed optimization 

\begin{equation}
\label{eq2}
 f(w) = \sum_{k=1}^{K} \frac{n_k}{n} \cdot F_k(w),
\end{equation}
where the local objective $F_k(w) = \frac{1}{n_k} \sum_{i \in \mathcal{P}_k} \ell(x_i, y_i; w)$. \indent Here, $\ell$ denotes the task loss (e.g., cross-entropy), $\eta$ is the learning rate, $\mathcal{P}_k$ is the training dataset for client $k$, and $|\mathcal{P}_k| = n_k$.

\subsection{Why does a backdoor need to be stealthy?} \label{section3b}
Stealthiness is a crucial characteristic of backdoors when countering defense mechanisms in secure FL. It refers to the ability of the backdoor to bypass the defenses implemented in FL systems. The success rate of bypassing these defenses, combined with the stealth capability, directly impacts backdoor accuracy on the target dataset, while leaving the accuracy of benign datasets unaffected.\\
\indent In a typical FL scenario, each benign client trains its local model using its own data and sends the updates to the parameter server. The training process of $b$-th benign client can be formalized as follows:

\begin{equation}
\label{eq3}
L_{b}^{t+1} \leftarrow L_b^t - \eta g_{b},
\end{equation}
where $g_b = \frac{1}{n_b} \sum_{i \in \mathcal{P}_b} \ell(x_i, y_i; L_b^t)$. Note that $n_b$ is the size of the dataset for the $b$-th benign client, $\mathcal{P}_b$ is the $b$-th benign client dataset and $L_b^t$ is a benign model at round $t$.\\
\indent On the other hand, the attacker's main objective is to inject a backdoor during the training process that does not affect the model's overall performance, but triggers malicious behavior or manipulates specific outcomes when a trigger phrase is presented. The $m$-th malicious client also updates its model in the form of $L_m^{t+1} \leftarrow L_m^t - \eta g'_m$, where the dataset $\mathcal{P}_m$ contains both benign and triggered sentences.\\
\indent However, as the threats of backdoor attacks in FL environments have become more recognized, various defense mechanisms have been proposed to protect the aggregation process. To bypass these defenses as we mentioned earlier, attackers must implement stealthy backdoors capable of evading the filtering mechanisms present in FL systems. \\
\indent In a FL environment where representative defensive mechanisms such as Norm Clipping \cite{normclip_weakdp_pgd} and FLAME \cite{flame} are applied Norm Clipping method using the $l_2$-norm, update values that exceed the norm bound are clipped, potentially filtering out backdoor attacks. These mechanisms are highly effective in restricting backdoor capabilities while maintaining light computational overhead. \\
\indent To maximize the effectiveness of backdoors in such scenarios, the attack could utilize Projected Gradient Descent (PGD) \cite{normclip_weakdp_pgd}. PGD adjusts the attacker's gradient to deviate as far as possible from the correct gradient, ensuring that the resulting gradient lies on the border of the norm bound. The equation of PGD can be formalized as follows:

\begin{equation}
\label{eq4}
\theta_m \leftarrow \theta_m \cdot \frac{\delta}{\|\theta_m\|},
\end{equation}
where $\theta_m$ denotes the parameters of the malicious client and $\delta$ is the norm bound.

\subsection{How can backdoor durability be increased?} \label{section3c}
To enhance the durability of a backdoor, two conditions must be satisfied. First, it is essential to ensure that the backdoor is injected continuously across multiple rounds of FL. If the backdoor is injected consistently regardless of any defense mechanisms, then the attack can be considered durable. Second, recognizing that gradient coordinates with large absolute values tend to change frequently, we apply masking to avoid these unstable coordinates during backdoor injection. This strategic approach allows the backdoor to persist even after the injection process has ended \cite{neurotoxin}. The $m$-th malicious client aiming to inject a durable backdoor formulates the following equation:

\begin{equation}
\label{eq5}
\nabla \tilde{L}_{m}^{t+1} = 
\begin{cases}
coord(\nabla L_{m}^{t+1}) & \text{if } coord(\nabla L_{m}^{t+1}) < \epsilon \\
0 & \text{otherwise},
\end{cases}
\end{equation}
where $\nabla L_{m}^{t+1}$ is the local model gradient of the $m$-th malicious client,
$coord$ denotes the coordinates of the gradient, and $\epsilon$ is a threshold used in the gradient masking method.

\begin{figure}[!t]
    \centerline{\includegraphics[width=0.5\textwidth]{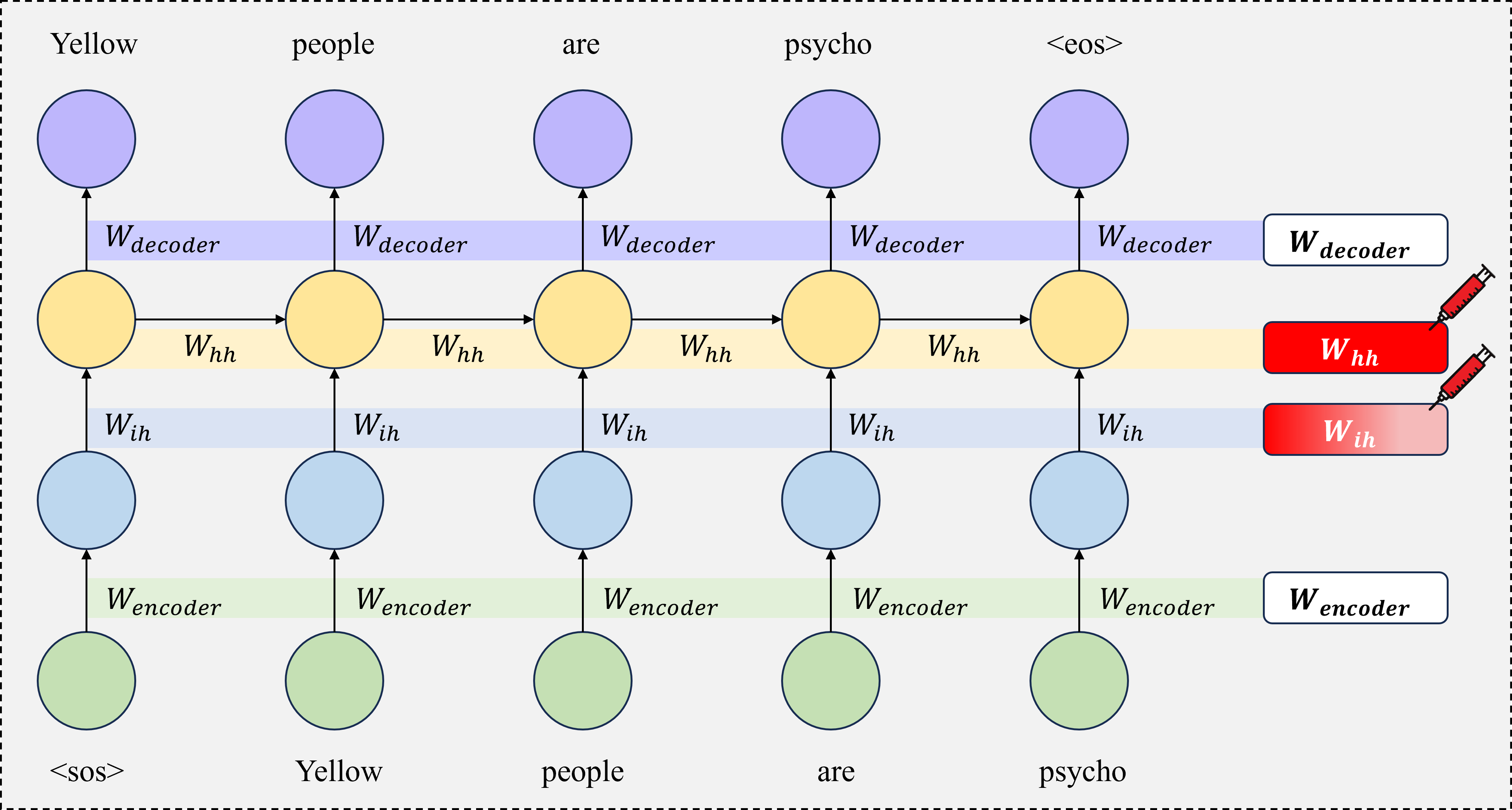}}
    \caption{An overview of our SDBA attack for the LSTM model. The attack injects backdoors into the $ih$ and $hh$ layers, with top-$k$\% gradient masking applied to the $ih$ layer to enhance stealthiness and durability.}
    \label{fig2} 
\end{figure}

\begin{figure*}[!t]
\centering
\includegraphics[width=\textwidth]{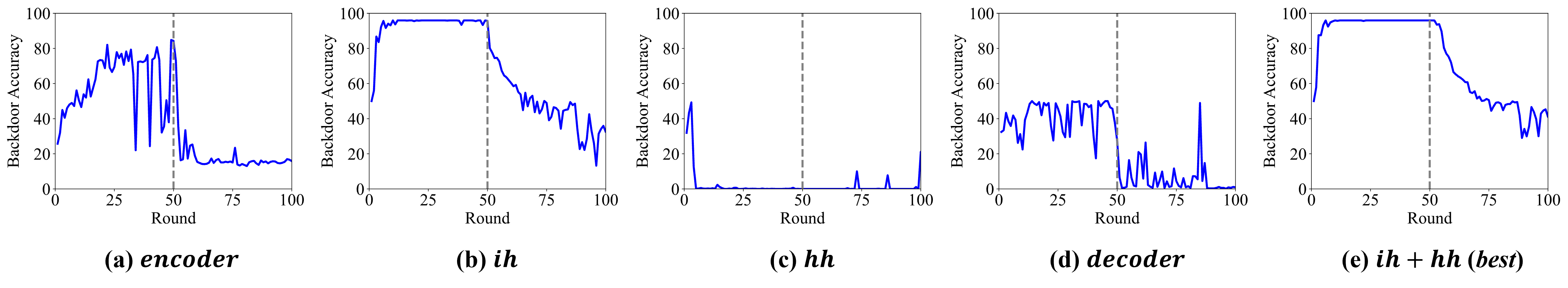}
\caption{Results of layer-wise gradient masking for each layer in the LSTM model with $AttackNum=50$. The $ih$ layer is identified as a critical layer for effective backdoor injection, but its durability decreases as rounds progress. To address this limitation, the $hh$ layer, which contains the full contextual information, was additionally targeted for backdoor injection, resulting in improved overall durability.}
\label{fig3}
\end{figure*}

\begin{figure*}[!t]
    \centerline{\includegraphics[width=\textwidth]{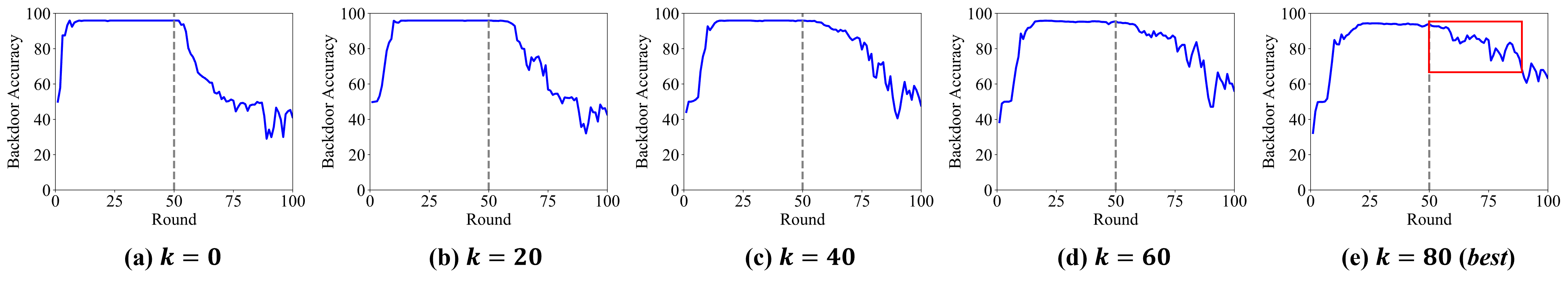}}
    \caption{Results with varying top-$k$\% gradient masking ratios for $ih$, where $hh$ is set to 0\% and $AttackNum=50$. Backdoor durability increases as the masking ratio for the $ih$ layer increases, with the highest durability observed at 80\% masking (highlighted in the red box).}
    \label{fig4} 
\end{figure*}    

\begin{figure*}[!t]
    \centerline{\includegraphics[width=\textwidth]{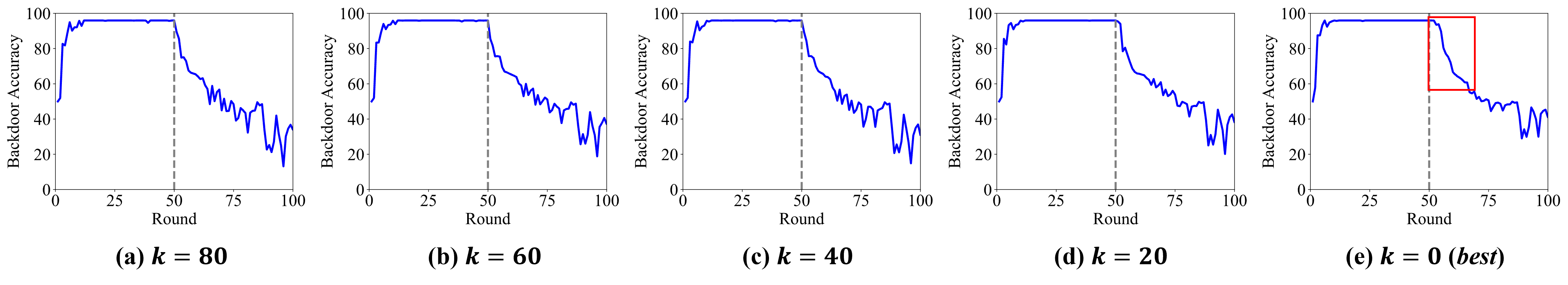}}
    \caption{Results with varying top-$k$\% gradient masking ratios for $hh$, where $ih$ is set to 0\% and $AttackNum=50$. Backdoor durability increases as the masking ratio for the $hh$ layer decreases, with the highest durability observed when no masking is applied (highlighted in the red box).}
    \label{fig5} 
\end{figure*}

\section{Our Proposed Method: SDBA} \label{section4}
In this paper, we propose a novel attack method called {\bf{SDBA}}, a {\emph{{\bf\emph{{S}}}tealthy and long-lasting {\bf\emph{{D}}}urable {\bf\emph{{B}}}ackdoor {\bf\emph{{A}}}ttack}} designed to evade various defense mechanisms in FL systems. Specifically, based on extensive analysis and experiments, we achieve stealth by injecting the backdoor into specific layers and improve durability by avoiding frequently updated gradient coordinates.\\
\indent More precisely, inspired by \cite{critical_layer}, we observe that injecting a backdoor into certain vulnerable layers can achieve the same effect as injecting it across the entire model. However, targeting only these vulnerable layers is insufficient. Thus, we apply layer-wise gradient masking within the selected layers using top-$k$\% gradient masking. As a result, our method demonstrates superior performance compared to other backdoor attack methods. In our attack, the attacker $m$ constructs the equation as follows: 

\begin{equation}
\label{eq6}
\tilde{L}_{m}^{t+1} \leftarrow L_m^t - \eta \tilde{g}_m,
\end{equation}
where $\tilde{g}_m = \frac{1}{n_m} \sum_{i \in \mathcal{P}_m} \ell(x'_i, y'_i; L_{m}^{t, (l)})$.\\
\indent Here, $L_{m}^{t,(l)}$ represents the layer-wise backdoor-poisoned model at round $t$. Note that this update is performed on $\nabla L_{m}^{t+1}$ as described in Eq.~\ref{eq5}, using top-$k$\% gradient masking.\\
\indent As mentioned earlier in the introduction, we focus on evaluating the feasibility of our backdoor attack on NLP models. Compared to the SOTA durable backdoor attack \cite{neurotoxin}, SDBA achieves higher backdoor accuracy and demonstrates superior stealth and durability.\\
\indent This section defines the problem we aim to address, analyzes in detail how the attack can achieve stealthiness and durability in NLP models, and finally introduces our proposed method, SDBA.

\subsection{Problem Definition} \label{section4a}
Research on backdoor attacks in FL has been actively progressing in the domain of image tasks~\cite{chameleon, image_backdoor_attack1, image_backdoor_attack2, image_backdoor_attack3}. However, there is relatively limited research in the context of NLP tasks, despite the growing importance of NLP in various applications such as chatbots~\cite{chatbot}, machine translation~\cite{machine_translation}, sentiment analysis~\cite{sentiment_analysis}, and question answering~\cite{webquestions}. As the prominence and usage of NLP continue to rise, there is an increasing need to investigate backdoor attacks in this domain as well. Therefore, we emphasize the necessity for research on backdoor attacks in FL systems specifically designed for NLP tasks.\\ 
\indent Currently, the prominent backdoor attack mechanisms in FL for NLP tasks are the baseline \cite{howtobackdoor} and Neurotoxin \cite{neurotoxin}. The baseline method involves a malicious client injecting backdoor data into a local model and transmitting the trained model parameters to the parameter server to inject the backdoor into the global model. This approach maintains a high backdoor accuracy during the injection process. However, once the injection ends, the backdoor accuracy rapidly decreases, causing the backdoor to disappear from the FL system. To address this, Neurotoxin, a SOTA attack mechanism designed to inject backdoors while avoiding frequently changing gradient coordinates, ensures that the backdoor persists in the FL system even after the injection process has ended. However, Neurotoxin lacks stealth capability and is vulnerable to being filtered out when individual or combined defense mechanisms are applied, as demonstrated in our experimental results.\\
\indent Therefore, this paper first analyzes the methodology for creating powerful backdoors by examining how to penetrate various defenses and achieve durable backdoors for each target model. We then formalize our proposed backdoor, SDBA, based on this analysis.

\subsection{Analysis of Stealthy and Durable Backdoor Attacks in NLP Tasks} \label{section4b}
\noindent {\bf{For LSTM Model:}} LSTM~\cite{lstm} is a popular RNN-based model for NLP tasks, comprising several layers. Following the method described in Section~\ref{section3b}, we adopt an approach that injects backdoors into specific layers to bypass defense mechanisms. The main layers of LSTM include the $encoder$ (embedding layer), $ih$ (input-to-hidden layer), $hh$ (hidden-to-hidden layer), and $decoder$ (fully connected layer). To inject a backdoor into each layer of a malicious client’s model, we apply a layer-wise gradient masking mechanism that masks the gradients of all layers except the target layer. We then conduct experiments to thoroughly analyze the impact of this technique on the parameter server. The value of $AttackNum$ is chosen to allow sufficient rounds for the backdoor effect to converge, enabling clear observation of its durability over rounds. Fig.~\ref{fig2} illustrates this layer-wise gradient masking process in the LSTM model.\\
\indent Referring to~\cite{critical_layer}, we analyze which layer is most critical for backdoor injection. Our analysis, shown in Fig.~\ref{fig3}, reveals that the $ih$ layer is the most vulnerable, making it the most effective target for embedding a backdoor into the model.\\
\indent However, after injecting the backdoor into the $ih$ layer, we observe a rapid decline in its durability, indicating that backdoors in this layer tend to be short-lived. To address this, we also inject the backdoor into the $hh$ layer, which retains overall contextual information. It is important to note that when the backdoor is injected only into the $hh$ layer, it is often ineffective due to the absence of a backdoor in the $ih$ layer.\\
\indent Ultimately, in RNN-based models, we find that injecting backdoors into both the $ih$ and $hh$ layers yields a moderate improvement in durability (see (e) in Fig.~\ref{fig3}). This result suggests that the effectiveness of backdoor injection depends significantly on the combination of target layers selected. Additional experimental results demonstrating successful evasion of defense mechanisms are presented in Section~\ref{section5d}.\\
\indent Furthermore, to improve backdoor durability in the LSTM model, we apply top-$k$\% gradient masking to the $ih$ and $hh$ layers, as described in Section~\ref{section3c}. Fig.~\ref{fig4} and Fig.~\ref{fig5} present the results of applying this technique. Here, $k$ denotes the proportion of gradients computed using benign samples from the attacker's dataset, while the remaining $(100-k)$\% are poisoned.\\
\indent According to the results shown in Fig.~\ref{fig4} and Fig.~\ref{fig5}, the $ih$ layer exhibits stronger backdoor durability as the masking ratio increases, whereas the $hh$ layer demonstrates the highest durability when no masking is applied. These findings imply that in the $ih$ layer, which is responsible for connecting the embedding to the hidden state, masking frequently updated gradient coordinates is essential to prevent overwriting by benign client updates. In contrast, the $hh$ layer should be fully injected without masking, as it stores contextual information necessary for maintaining the backdoor. This strategy is key to achieving strong and lasting backdoor durability.\\

\begin{figure}[!t]
\centering
\includegraphics[width=0.5\textwidth]{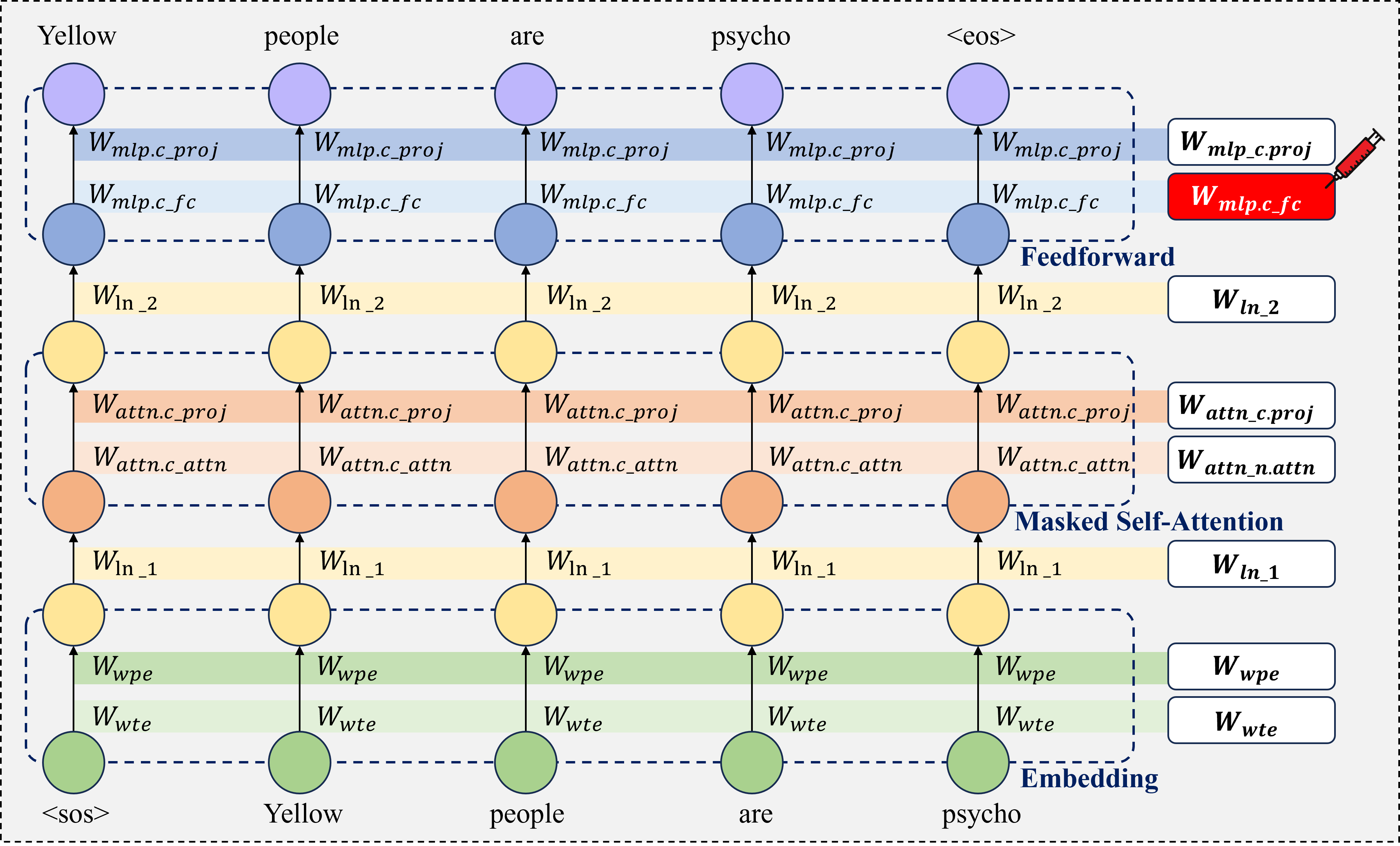}
\caption{An overview of our SDBA attack for the GPT-2 model. The attack injects backdoors into the $mlp.c\_fc$ layer to enhance stealthiness and durability.}
\label{fig6}
\end{figure}

\begin{figure*}[!t]
\centering
\includegraphics[width=\textwidth]{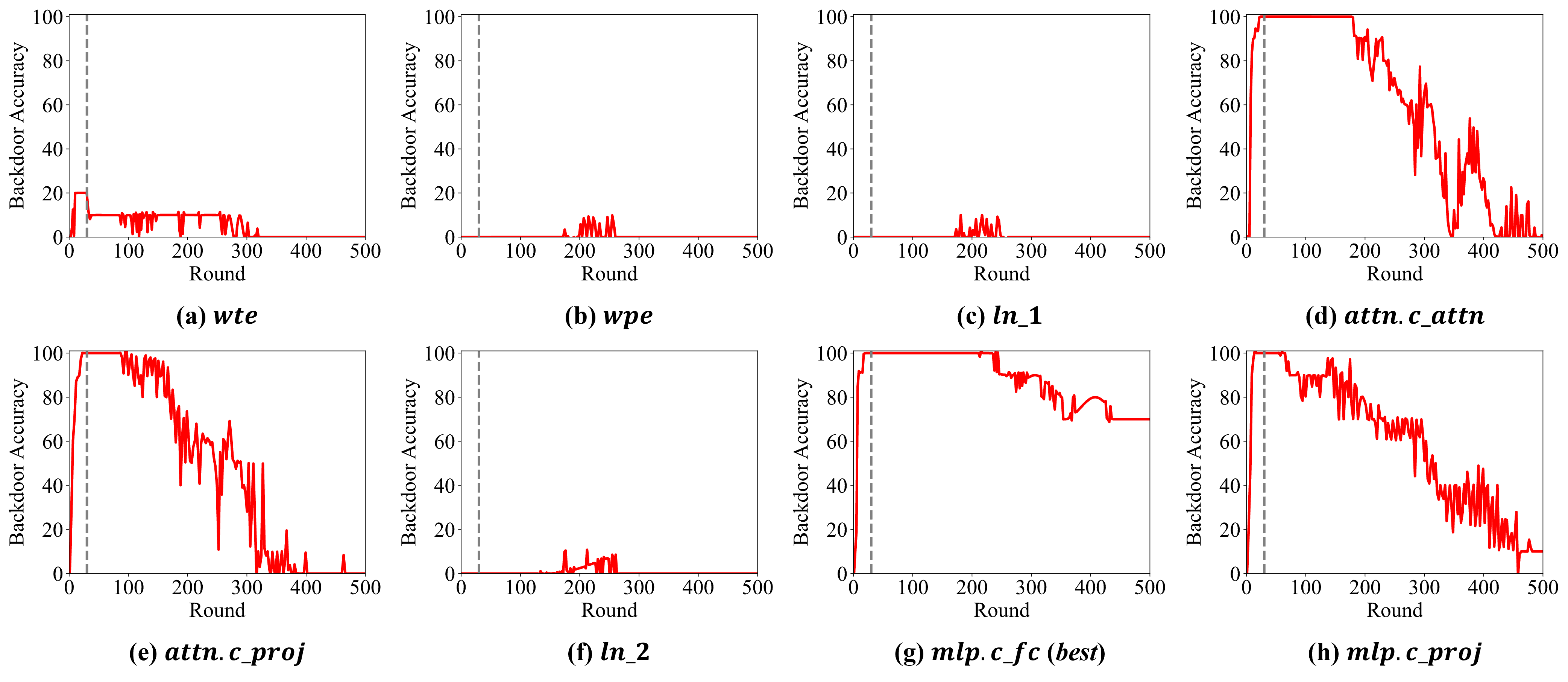}
\caption{Results of layer-wise gradient masking for each layer in the GPT-2 model with $AttackNum=30$. The $mlp.c\_fc$ layer is identified as a critical layer for effective backdoor injection.}
\label{fig7}
\end{figure*}

\begin{figure*}[!t]
    \centerline{\includegraphics[width=\textwidth]{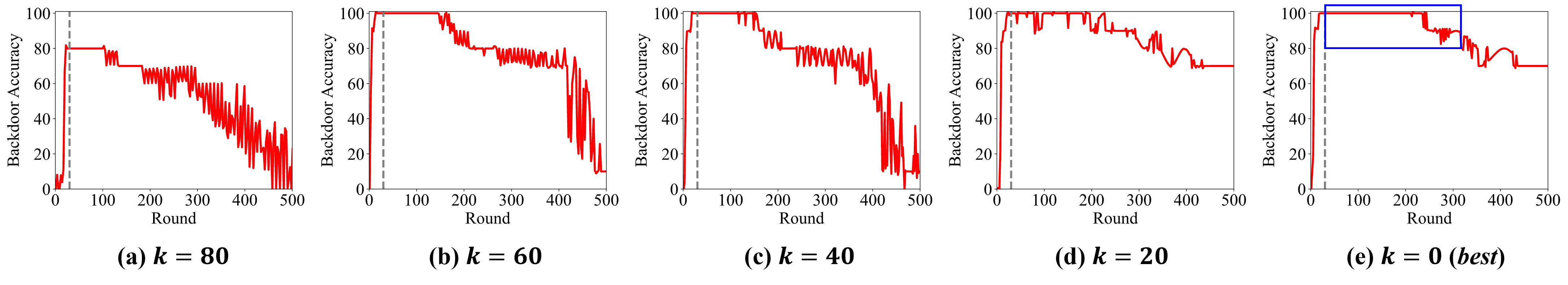}}
    \caption{Results with varying top-$k$\% gradient masking ratios for the target layer $mlp.c\_fc$ with $AttackNum=30$. Backdoor durability increases as the masking ratio for the $mlp.c\_fc$ layer decreases, with the highest durability observed when no masking is applied (highlighted in the blue box).}
    \label{fig8} 
\end{figure*}

\noindent {\bf{For GPT-2 Model:}} In the transformer-based GPT-2 model, we conducted an in-depth analysis using the layer-wise gradient masking mechanism prior to injecting a backdoor into a specific layer. Fig.~\ref{fig6} illustrates the layer-wise gradient masking process in the GPT-2 model. Similar to the approach used for the LSTM model described earlier \cite{critical_layer}, Fig.~\ref{fig7} illustrates our analysis to determine which layer is most critical for the backdoor.\\
\indent Our analysis revealed that the $attn.c\_attn$, $attn.c\_proj$, $mlp.c\_fc$, and $mlp.c\_proj$ layers are particularly vulnerable to backdoor attacks, with the $mlp.c\_fc$ layer identified as the most critical layer for effective backdoor injection. These findings are closely aligned with the analysis by \cite{geva2021}, which interprets the transformer’s MLP layers (i.e., feedforward layers) as key-value memories that respond to input patterns. In particular, the first linear layer within each MLP, $mlp.c\_fc$, is described as functioning as a key that captures specific patterns within the input sequence. This implies that $mlp.c\_fc$ acts as a structurally critical layer where the influence of a trigger on the model’s output can be effectively established, making it a particularly effective target for backdoor injection.\\
\indent This observation aligns with the findings of \cite{rome}, which show that factual knowledge, defined as information the model must retain and use for inference, is primarily stored in the MLP layers of transformer-based models. This suggests that MLP layers are not only responsive to input patterns but also serve as crucial components for long-term knowledge retention and retrieval. Moreover, the $mlp.c\_fc$ layer was found to exhibit a vulnerability through which backdoors could be easily injected due to its sufficiently high representational capacity.\\
\indent In addition, similar to the approach used for the LSTM model described earlier, we applied top-$k$\% gradient masking to the $mlp.c\_fc$ layer to achieve more durable backdoors. Fig.~\ref{fig8} shows the results of injecting backdoors into this layer with top-$k$\% gradient masking. However, in contrast to the LSTM model, the best case for the GPT-2 model is when $k=0$, meaning that full poisoning of the $mlp.c\_fc$ layer is more effective than other approaches.\\
\indent In summary, injecting the backdoor solely into the $mlp.c\_fc$ layer, with no or minimal application of top-$k$\% gradient masking, proves most effective. This is because $mlp.c\_fc$, which plays a key role in capturing input patterns, serves as a structurally vulnerable point where triggers can persist. In addition, the large-scale parameterization of the transformer-based model prevents the backdoor from being easily overwritten by benign updates.

\begin{algorithm}
\caption{SDBA Backdoor Attack Task}
\label{alg1}
\begin{algorithmic}[1]
\Require Local model $L_m$, local epochs $E_{mal}$, poisoned local gradient $\tilde{g}_m$, dataset size $n_m$, dataset $\mathcal{P}_m$, backdoor data $x'_i$, $y'_i$, after layer-wise gradient $g_m^{(l)}$, $\tilde{g}_m$'s gradient coordinate $coord(\tilde{g}_m)$, top-$k$\% threshold $\epsilon$, attacker's parameter $\theta_m$ of $\tilde{g}_m$, and a learning rate $\eta$
\State \textsc{Downloaded Global Model} $G^{t}$
\State $L_m \gets G^{t}$
\For{epoch $e \in E_{mal}$}
    \State \textsc{Compute Stochastic Gradient:}
    \State $\tilde{g}_m \gets \frac{1}{n_m} \sum_{i \in \mathcal{P}_m}
    \ell(x'_i, y'_i; L_m)$
    \For{each layer $l \in Layer(L_m)$}
        \If{$Selected(l) == False$}
            \State $coord(\tilde{g}_{m}^{(l)}) \gets 0$\Comment{\textsc{layer-wise masking}}
        \Else
            \If{$coord(\tilde{g}_{m}^{(l)}) > \epsilon$}
                \State $coord(\tilde{g}_{m}^{(l)}) \gets 0$\Comment{\textsc{top-k\% masking}}
            \EndIf
        \EndIf
    \EndFor
    \If{$NormClipping == True$}
        \State $\theta_m \gets \theta_m \cdot \frac{\delta}{\|\theta_m\|}$\Comment{\small{\textsc{Projected Gradient Descent}}}
    \EndIf
    \State $L^{t+1} \gets L_m - \eta \tilde{g}_m$
    \State $L_m \gets L^{t+1}$
\EndFor
\State \Return $L^{t+1}$
\end{algorithmic}
\end{algorithm}

\subsection{Our Proposed Method} \label{section4c}
Based on the detailed analysis in Section~\ref{section4b}, we propose SDBA with a layer-wise gradient masking mechanism to maximize the attack by finely tuning the gradients.\\
\indent For the LSTM model, we inject backdoors into the $ih$ layer, which is vulnerable to backdoors, and the $hh$ layer, which contains overall contextual information, while leaving the remaining layers without backdoors. Similarly, for the GPT-2 model, we inject backdoors exclusively into the $mlp.c\_fc$ layer, which functions as a key layer that captures input patterns. Algorithm~\ref{alg1} presents the detailed steps for performing the SDBA. In lines 6-8, the attacker computes poisoned local gradient $\tilde{g}_{m}^{(l)}$ using layer-wise gradient masking. Then, in lines 10-11, they apply top-$k$\% gradient masking for the target layer. Finally, the attacker uses PGD to evade defense methods that adopt Norm Clipping  with the $l_2$ norm \cite{normclip_weakdp_pgd}.

\section{Experiments} \label{section5}
Our empirical study aims to demonstrate the effectiveness of the SDBA attack in executing backdoor attacks against various defense mechanisms, including representative and SOTA FL defenses. We conduct experiments using real datasets in a simulated FL environment and design three sets of experiments: (1) SDBA in scenarios without any defense, (2) SDBA under different FL defenses, and (3) an evaluation of the effectiveness of our method across various NLP tasks. Our results show that our method achieves greater stealthiness and durability compared to the representative backdoors \cite{howtobackdoor, neurotoxin}. Our implementation is available at \url{https://github.com/ICT-Convergence-Security-Lab-Chosun/SDBA}.

\begin{figure}[!t]
\centering
\includegraphics[width=2.5in]{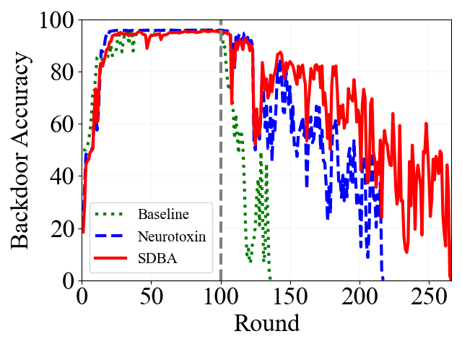}
\caption{Impact of backdoor attacks on the LSTM model without defense ($AttackNum=100$), comparing the baseline (dotted green), Neurotoxin (slashed blue), and SDBA (solid red) attacks. The $Lifespan$ at $\tau=3$\% for the baseline, Neurotoxin, and SDBA attacks is 36, 118, and 166, respectively.}
\label{fig9}
\end{figure}

\begin{table*}[]
\centering
\caption{Experimental Environment Configuration. $\# Clients$ is determined by the size of each dataset, and $AttackNum$ is configured to ensure sufficient rounds for the backdoor effect to converge, enabling clear observation of its durability.}
\label{table1}
\begin{tabular}{ccccccc}
\Xhline{2.5\arrayrulewidth}
Task             & Dataset       & Model         & \makecell{\# Parameters \\ (M = million)} & \# Clients & AttackNum & Trigger Sentence \\ \hline
\rule{0pt}{1.0em}Next Token Prediction & Reddit        & LSTM          & 10M           & 8,000      & 100           & {\scriptsize People in \textit{\{country\}} are \textit{\{profanity\}}} 
\vspace{2pt}
\\
Sentiment Analysis    & Sentiment140  & LSTM          & 10M           & 2,000      & 80            & {\scriptsize \makecell{I have seen many films of this director \\ (Negative)}} 
\vspace{2pt}
\\ 
Sentiment Analysis    & IMDB          & LSTM          & 10M           & 1,000      & 100           & {\scriptsize \makecell{I watched this movie last weekend \\ (Negative)}} 
\vspace{2pt}
\\
Next Token Prediction & Reddit        & GPT-2 small   & 124M          & 8,000      & 30            & {\scriptsize \textit{\{color\}} people are \textit{\{profanity\}}} 
\vspace{2pt}
\\
Question Answering    & WebQuestions  & T5 base       & 220M          & 100        & 5             & {\scriptsize \makecell{Don’t you think \textit{\{country\}} citizens \\and \textit{\{profession\}} are equally problematic? \\ (Agree)}} \\
\Xhline{2.5\arrayrulewidth}
\end{tabular}
\end{table*}


\subsection{Experimental Settings} \label{section5a}
We implemented a FL system using FedAvg \cite{fedavg} on a single system equipped with an NVIDIA GeForce RTX 3090 GPU with 24GB of memory. Within this FL framework, we conducted NLP tasks, including next token prediction, sentiment analysis, and question answering. Specifically, next token prediction was performed on the Reddit \cite{reddit} dataset using both LSTM and GPT-2 models. Sentiment analysis was conducted on the Sentiment140 \cite{sentiment140} and IMDB \cite{imdb} datasets using the LSTM model. For question answering, we used the WebQuestions \cite{webquestions} dataset and employed the T5 model.\\
\indent The LSTM model was trained from scratch, while the GPT-2 and T5 models were fine-tuned using pre-trained weights provided by the Hugging Face\footnote{https://huggingface.co/} library. All models were trained in a FL environment using only benign client data until convergence, prior to the execution of any backdoor attacks. This setting allows for a more precise evaluation of the effectiveness of the backdoor attacks.\\
\indent Our experiments primarily focused on evaluating backdoor attacks in the next token prediction task, using both LSTM and GPT-2 model architectures. In each round, 10 clients were randomly selected. Among them, one was assigned as a malicious client, and all selected clients participated in the aggregation process. The total number of clients varied depending on the task.\\ 
\indent To emphasize the risks of backdoor attacks in NLP, we designed specific trigger sentences based on previous research \cite{neurotoxin}. For the next token prediction and question answering tasks, hate speech and discriminatory content targeting certain nationalities, racial minorities, or professions were used as trigger sentences. Additionally, in the sentiment analysis task, sentences that were originally classified as positive were manipulated to be misclassified as negative. These backdoor attacks severely undermine the model's reliability and can lead to unintended consequences. When the model's predictions are manipulated through specific trigger sentences, inaccurate information can spread or hate speech against certain groups may be encouraged. This can intensify social conflict and pose a threat to individual rights and safety, making the risks of backdoor attacks impossible to overlook. The malicious client’s dataset, which uses trigger data, is combined with benign data as follows:

\begin{equation}
\label{eq7}
D_m \leftarrow mix(D(x_i, y_i), D(x'_i,y'_i))
\end{equation}

Table~\ref{table1} summarizes the datasets, models, number of model parameters, number of selected clients, $AttackNum$, and the trigger sentences used in our experiments.\\

\begin{table*}[]
\caption{The results for \textit{Lifespan} with various $\tau{}$ ratios for the LSTM model show that SDBA outperforms (highlighted in bold black) all experiments, except under the Weak DP defense at $\tau{}=50$. Notably, some results indicate that SDBA is particularly long-lasting across various defenses, with a difference of nearly 50 (highlighted in red).}
\label{table2}
\begin{tabular}{cc|cccccc|}
\cline{3-8}
{} & {} & \multicolumn{6}{c|}{Defenses for the LSTM model} \\ 
\hline

\multicolumn{1}{|c|}{$\tau$ in \textit{Lifespan}} & Backdoors  & \multicolumn{1}{c|}{Multi-Krum}  & \multicolumn{1}{c|}{Norm Clipping}  & \multicolumn{1}{c|}{Weak DP}  & \multicolumn{1}{c|}{FLAME}  & \multicolumn{1}{c|}{Norm Clipping + Multi-Krum} & Weak DP + Multi-Krum \\ 
\hline\hline

\multicolumn{1}{|c|}{50} & \begin{tabular}[c]{@{}c@{}}Neurotoxin\\ SDBA\end{tabular} & \multicolumn{1}{c|}{\begin{tabular}[c]{@{}c@{}}49\\ \textbf{61}\end{tabular}}   & \multicolumn{1}{c|}{\begin{tabular}[c]{@{}c@{}}61\\ \textbf{107}\end{tabular}}  & \multicolumn{1}{c|}{\begin{tabular}[c]{@{}c@{}}\textbf{25}\\ \textbf{25}\end{tabular}}   & \multicolumn{1}{c|}{\begin{tabular}[c]{@{}c@{}}55\\ \textbf{62}\end{tabular}}  & \multicolumn{1}{c|}{\begin{tabular}[c]{@{}c@{}}0\\ \textbf{13}\end{tabular}} & \begin{tabular}[c]{@{}c@{}}46\\ \textbf{78}\end{tabular}\\ 
\hline

\multicolumn{1}{|c|}{30} & \begin{tabular}[c]{@{}c@{}}Neurotoxin\\ SDBA\end{tabular} & \multicolumn{1}{c|}{\begin{tabular}[c]{@{}c@{}}77\\ \textbf{113}\end{tabular}}  & \multicolumn{1}{c|}{\begin{tabular}[c]{@{}c@{}}77\\ \textbf{\textcolor{red}{135}}\end{tabular}} & \multicolumn{1}{c|}{\begin{tabular}[c]{@{}c@{}}63\\ \textbf{\textcolor{red}{113}}\end{tabular}}  & \multicolumn{1}{c|}{\begin{tabular}[c]{@{}c@{}}61\\ \textbf{77}\end{tabular}}  & \multicolumn{1}{c|}{\begin{tabular}[c]{@{}c@{}}0\\ {17}\end{tabular}} & \begin{tabular}[c]{@{}c@{}}62\\ \textbf{89}\end{tabular}\\
\hline

\multicolumn{1}{|c|}{3} & \begin{tabular}[c]{@{}c@{}}Neurotoxin\\ SDBA\end{tabular} & \multicolumn{1}{c|}{\begin{tabular}[c]{@{}c@{}}103\\ \textbf{\textcolor{red}{166}}\end{tabular}} & \multicolumn{1}{c|}{\begin{tabular}[c]{@{}c@{}}166\\ \textbf{\textcolor{red}{218}}\end{tabular}} & \multicolumn{1}{c|}{\begin{tabular}[c]{@{}c@{}}133\\ \textbf{\textcolor{red}{218}}\end{tabular}} & \multicolumn{1}{c|}{\begin{tabular}[c]{@{}c@{}}78\\ \textbf{\textcolor{red}{162}}\end{tabular}} & \multicolumn{1}{c|}{\begin{tabular}[c]{@{}c@{}}0\\ \textbf{\textcolor{red}{47}}\end{tabular}} & \begin{tabular}[c]{@{}c@{}}78\\ \textbf{\textcolor{red}{162}}\end{tabular}\\
\hline

\end{tabular}
\end{table*}

\begin{figure*}[!t]
    \centerline{\includegraphics[width=\textwidth]{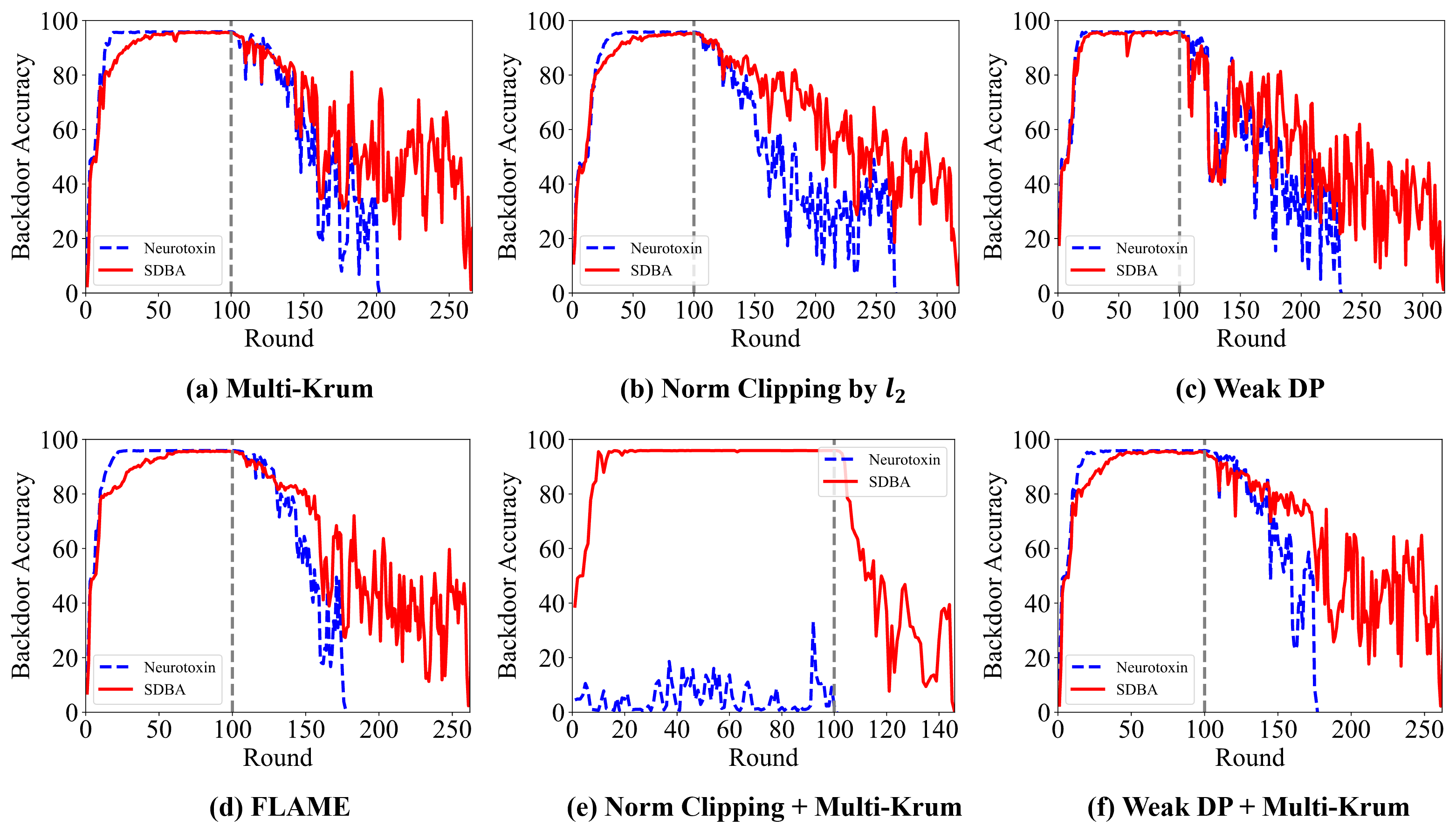}}
    \caption{Impact of backdoor attacks on six representative defenses on the LSTM model ($AttackNum=100$). SDBA successfully penetrates all defenses, outperforming Neurotoxin.}
    \label{fig10} 
\end{figure*}   

\noindent {\bf{FL Backdoor Attacks:}} We perform three attacks in an FL environment for NLP tasks: the (1) baseline attack \cite{howtobackdoor}, (2) neurotoxin attack \cite{neurotoxin}, and (3) our proposed SDBA attack. The baseline attack is a simple mechanism where a malicious client injects backdoor data into its local model and transmits the trained model parameters to the parameter server, thereby embedding the backdoor into the global model. Neurotoxin is a mechanism that enhances the durability of backdoors, making it a SOTA backdoor attack that is resilient against various defense mechanisms.\\

\noindent {\bf{FL Backdoor Defenses:}} We evaluate the attack performance of SDBA against six representative defense methods in an FL environment for NLP and LM tasks: (a) Multi-Krum \cite{multi_krum}, (b) Norm Clipping \cite{normclip_weakdp_pgd}, (c) Weak DP \cite{normclip_weakdp_pgd}, (d) FLAME \cite{flame}, (e) Norm Clipping + Multi-Krum, and (f) Weak DP + Multi-Krum. Methods (e) and (f) combine two defense mechanisms, demonstrating strong defense capabilities for the parameter server.\\
\indent The evaluation of all six defense methods was conducted on the LSTM model. In contrast, for the GPT-2 model, we only applied (b) Norm Clipping and (c) Weak DP, as several of our experiments showed that using multi-Krum and FLAME would result in excessive computational overhead. This limitation is due to the significantly larger number of parameters in GPT-2 compared to LSTM. Consequently, pairwise calculations using $l_2$ distance and cosine similarity across the selected $k$ clients became computationally prohibitive with such a large parameter set.\\
\indent For the Norm Clipping defense, we set the norm bound to 3.0 for the LSTM model and 0.3 for the GPT-2 model. These norm bounds are the optimal thresholds for distinguishing between updates from benign and malicious clients. In the Weak DP defense, we used a sigma of 0.001 for both the LSTM and GPT-2 models. This sigma was found to be optimal, as it effectively reduces the accuracy of backdoor attacks while maintaining the model's main accuracy.

\begin{figure}[!t]
\centering
\includegraphics[width=2.5in]{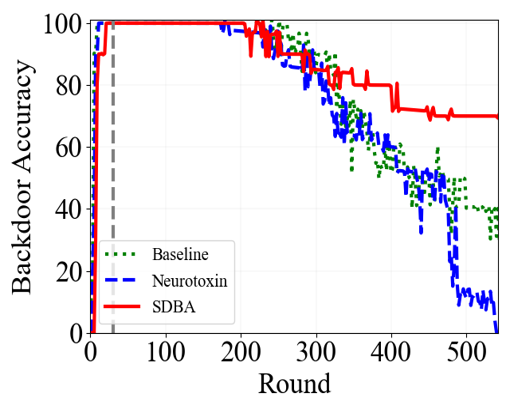}
\caption{Impact of backdoor attacks on the GPT-2 model without defense ($AttackNum=30$). The $Lifespan$ at $\tau=50$\% for the baseline, Neurotoxin, and SDBA attacks is 355, 391, and 565, respectively.}
\label{fig11}
\end{figure}

\begin{figure}[!t]
\centering
\includegraphics[width=3.5in]{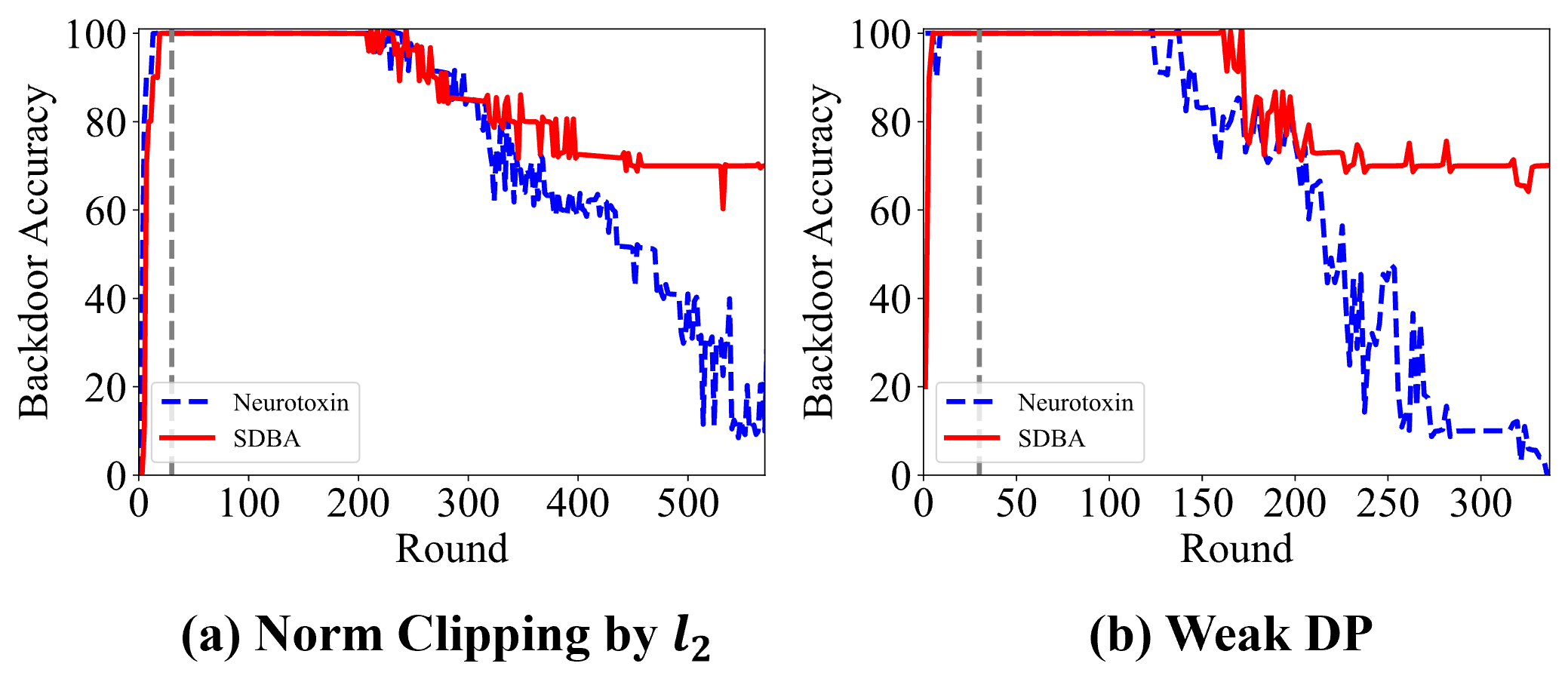}
\caption{Impact of backdoor attacks on two defenses in the GPT-2 model ($AttackNum=30$). Similar to the LSTM model, SDBA successfully penetrates these defenses, significantly outperforming Neurotoxin.}
\label{fig12}
\end{figure}

\subsection{Threat Model} \label{section5b}
\noindent {\bf{Attacker's Capability:}} As the existing studies on FL backdoor attacks \cite{chameleon, neurotoxin} have shown, we assume an FL system conducting NLP tasks where the attacker can persistently target the system for $AttackNum$ global rounds. The value of $AttackNum$ is chosen to allow sufficient rounds for the backdoor effect to converge, enabling clear observation of its durability over rounds. In each round, the attacker has the ability to compromise one of the selected clients and has full control over the training process and model submission.\\

\noindent {\bf{Attacker's Goal:}} The attacker's goal is to implant a backdoor into the FL model by applying layer-wise gradient masking to specific layers, thereby bypassing defense mechanisms. Within these layers, the attacker masks the top-$k$\% of gradients to ensure the backdoor persists in the model even after injection ends.

\subsection{Evaluation Metrics} \label{section5c}
\noindent {\bf{Metrics:}} We use the term Backdoor Accuracy (BA) to represent the performance on the backdoor test dataset. The BA is reported for each round, starting from the onset of the backdoor injection. Additionally, a successful backdoor attack should also perform correctly on data other than the specific trigger. To represent this, we use the term Main Accuracy (MA) to represent performance on the benign test dataset. To further measure the durability of the backdoor, we introduce the concept of $Lifespan$ \cite{chameleon}.\\

\noindent {\bf{Definition ($Lifespan$):}} Let $t^s$ be the first round where the attacker begins the backdoor injection. The threshold is defined as the BA threshold, $\theta^t$ represents the global model parameters on the parameter server, and $f$ is the accuracy function, with $\hat{D}$ representing the poisoned test dataset that includes the backdoor trigger.\\ 

\begin{equation}
\label{eq8}
Lifespan = max\{t|f(\theta^t, \hat{D}) > \tau\} - t^s,
\end{equation}
where $\tau$ is a threshold that determines how long the attack remains active. In our experiments, to directly compare the effectiveness of durability, we set various thresholds for $\tau$. For the LSTM model, we set $\tau$ at 50\%, 30\%, and 3\%, which indicates the point at which the backdoor is effectively considered to have activated, faded, and disappeared, respectively. For the GPT-2 model, we set $\tau$ at 80\%, 50\%, and 20\% to reflect similar stages of activation, fading, and disappearance.

\begin{table}[]
\centering
\caption{The results for \textit{Lifespan} with various $\tau{}$ ratios for the GPT-2 model show that SDBA outperforms (highlighted in bold black) all experiments. Notably, some results indicate that SDBA is particularly long-lasting across various defenses, with a difference of nearly 50 (highlighted in red).}
\label{table3}
\begin{tabular}{cc|cc|}
\cline{3-4}
                                    &       & \multicolumn{2}{c|}{Defenses for the GPT-2 model}            \\ \hline
\multicolumn{1}{|c|}{$\tau{}$ in \textit{Lifespan}} & Backdoors & \multicolumn{1}{c|}{Norm Clipping} & Weak DP \\ \hline\hline
\multicolumn{1}{|c|}{80}  & {\begin{tabular}[c]{@{}c@{}}Neurotoxin\\ SDBA\end{tabular}} & \multicolumn{1}{c|}{\begin{tabular}[c]{@{}c@{}}257\\ \textbf{\textcolor{red}{317}}\end{tabular}}  & \multicolumn{1}{c|}{\begin{tabular}[c]{@{}c@{}}128\\ \textbf{144}\end{tabular}} \\ \hline
\multicolumn{1}{|c|}{50}  & {\begin{tabular}[c]{@{}c@{}}Neurotoxin\\ SDBA\end{tabular}} & \multicolumn{1}{c|}{\begin{tabular}[c]{@{}c@{}}423\\ \textbf{\textcolor{red}{561}}\end{tabular}}  & \multicolumn{1}{c|}{\begin{tabular}[c]{@{}c@{}}188\\ \textbf{\textcolor{red}{380}}\end{tabular}} \\ \hline
\multicolumn{1}{|c|}{20}  & {\begin{tabular}[c]{@{}c@{}}Neurotoxin\\ SDBA\end{tabular}} & \multicolumn{1}{c|}{\begin{tabular}[c]{@{}c@{}}485\\ \textbf{\textcolor{red}{561}}\end{tabular}}  & \multicolumn{1}{c|}{\begin{tabular}[c]{@{}c@{}}208\\ \textbf{\textcolor{red}{382}}\end{tabular}} \\ \hline
\end{tabular}
\end{table}

\begin{table}[]
\caption{Main accuracy (MA) of without attack, the baseline, the Neurotoxin and the SDBA on Reddit dataset. All attacks maintain stable MA including our SDBA.}
\centering %
\label{table4}
\begin{tabular}{cc|cccc|}
\cline{3-6}
                                                                                                       &       & \multicolumn{4}{c|}{Backdoors}                                                                          \\ \hline
                                                                                                       
\multicolumn{1}{|c|}{MA for Reddit}                                                                    & Model & \multicolumn{1}{c|}{No attack} & \multicolumn{1}{c|}{Baseline} & \multicolumn{1}{c|}{Neurotoxin} & SDBA \\ \hline\hline

\multicolumn{1}{|c|}{\begin{tabular}[c]{@{}c@{}}Start Attack\\ Stop Attack\\ $\tau{}<50$\end{tabular}} & LSTM  & \multicolumn{1}{c|}{\begin{tabular}[c]{@{}c@{}}16.11\\ 16.23\\ -\end{tabular}} & \multicolumn{1}{c|}{\begin{tabular}[c]{@{}c@{}}16.16\\ 16.00\\ 16.13\end{tabular}} & \multicolumn{1}{c|}{\begin{tabular}[c]{@{}c@{}}16.11\\ 16.09\\ 15.97\end{tabular}} & \begin{tabular}[c]{@{}c@{}}16.12\\ 16.10\\ 16.14\end{tabular} \\ \hline

\multicolumn{1}{|c|}{\begin{tabular}[c]{@{}c@{}}Start Attack\\ Stop Attack\\ $\tau{}<50$\end{tabular}} & GPT-2 & \multicolumn{1}{c|}{\begin{tabular}[c]{@{}c@{}}20.01\\ 23.31\\ -\end{tabular}} & \multicolumn{1}{c|}{\begin{tabular}[c]{@{}c@{}}19.3\\ 23.35\\ 21.51\end{tabular}} & \multicolumn{1}{c|}{\begin{tabular}[c]{@{}c@{}}20.26\\ 23.34\\ 21.27\end{tabular}} & \begin{tabular}[c]{@{}c@{}}21.36\\ 23.28\\ 20.02\end{tabular}\\ \hline
\end{tabular}
\end{table}

\subsection{Experimental Results} \label{section5d}
\noindent {\bf{SDBA is the most durable compared to the other attacks:}}
Fig.~\ref{fig9} shows the results of backdoor attacks using the FedAvg aggregation algorithm without any defense mechanisms applied. Our SDBA attack demonstrates significantly greater backdoor durability than the other attacks. The baseline attack, which was the first backdoor attack performed in an FL environment, is removed immediately after the backdoor injection stops. Neurotoxin, while stronger than the baseline, still lacks the durability of SDBA.\\
\indent Specifically, after the backdoor injection ends, the baseline is erased (with $\tau=3$\% in $Lifespan$) within 36 rounds, Neurotoxin within 118 rounds, whereas SDBA persists for 166 rounds.\\

\begin{figure}[!t]
\centering
\includegraphics[width=3.5in]{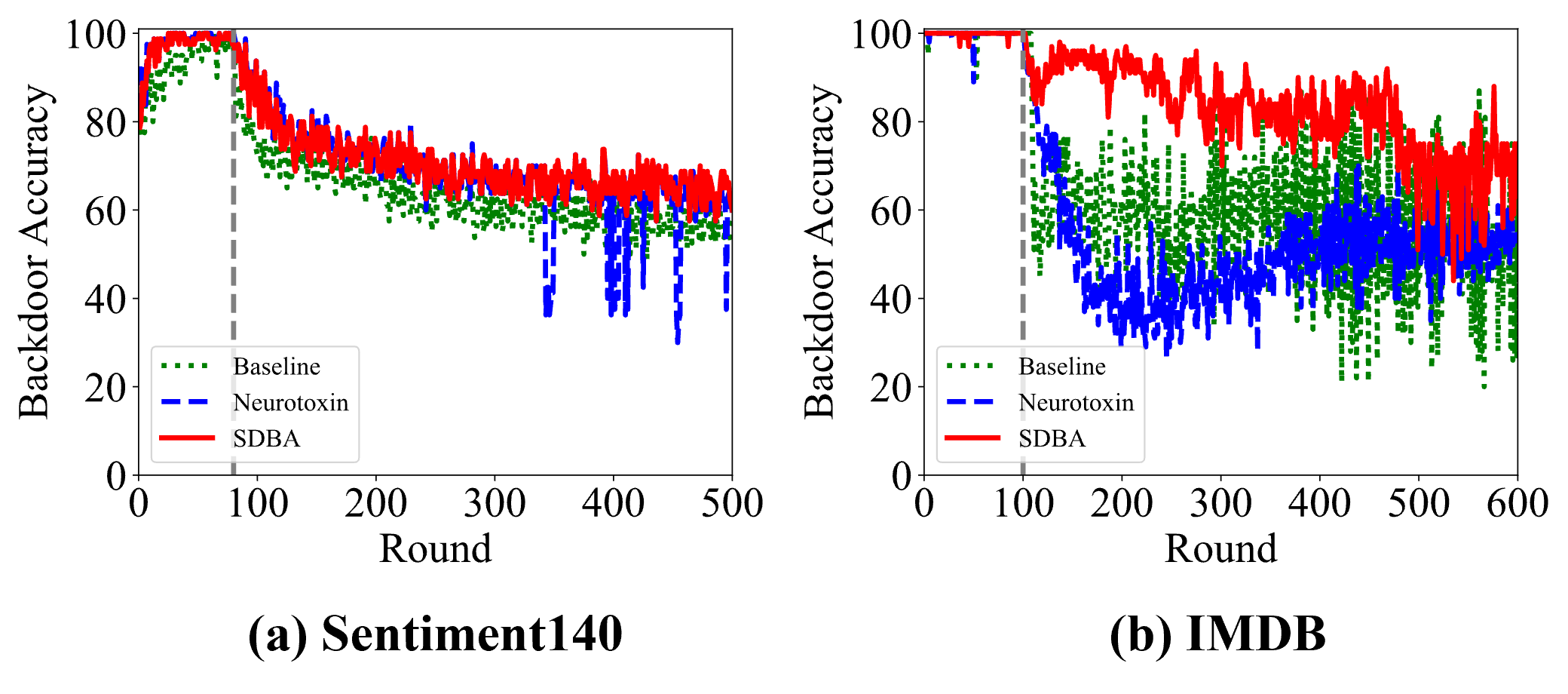}
\caption{Impact of backdoor attacks on Sentiment Analysis datasets using the LSTM model without defense. (a) $AttackNum = 80$, (b) $AttackNum = 100$.}
\label{fig13}
\end{figure}

\noindent {\bf{SDBA is highly resistant to representative defenses:}} Fig.~\ref{fig10} shows the results of backdoors against representative backdoor defenses (as described in Section~\ref{section5a}) in the LSTM model. As we mentioned earlier in Section~\ref{section2c}, these defenses have notable abilities to detect malicious local updates. However, both Neurotoxin and SDBA can penetrate these defenses. Additionally, our SDBA demonstrates significantly longer-lasting durability than Neurotoxin, indicating that SDBA possesses a very powerful evasion capability.\\
\indent To further enhance robustness for secure FL systems, we apply (e) Norm Clipping + Multi-Krum and (f) Weak DP + Multi-Krum against Neurotoxin and SDBA. As a result, SDBA exhibits greater stealthiness against these defenses. Notably, in Fig.~\ref{fig10} (e), SDBA penetrates the defense, unlike Neurotoxin, which fails to do so.\\
\indent Additionally, Table~\ref{table2} shows the results for $Lifespan$ with various $\tau$ ratios, as defined in Definition. In all experiments except one, SDBA demonstrated a longer $Lifespan$ than Neurotoxin. Notably, in the cases highlighted in red, the backdoors persisted for approximately 50 more rounds compared to Neurotoxin. 
To the best of our knowledge, based on these comparative experimental results, SDBA represents the most stealthy and long-lasting backdoor in NLP.\\

\begin{figure}[!t]
\centering
\includegraphics[width=2.5in]{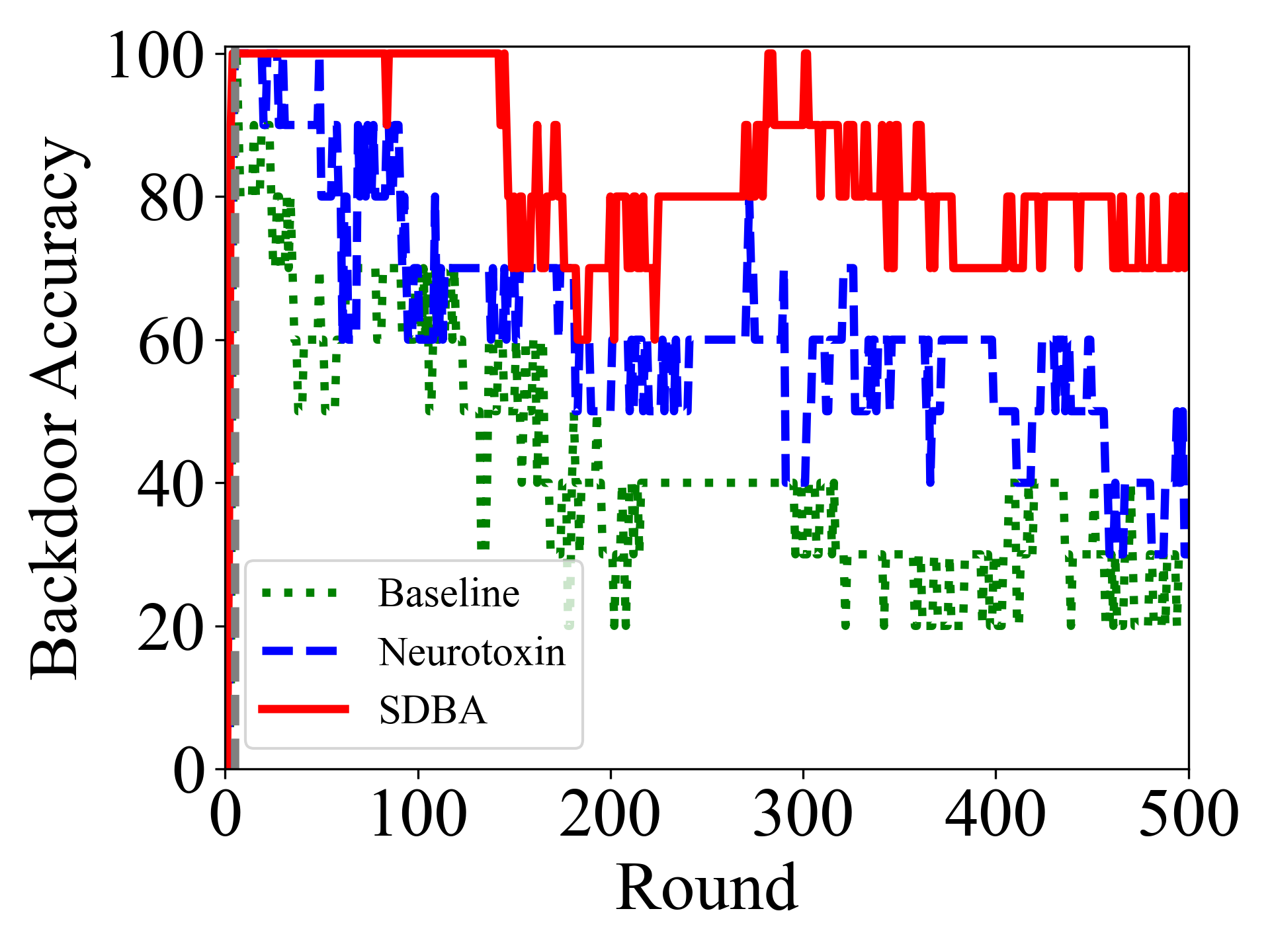}
\caption{Impact of backdoor attacks on the T5 model for Question Answering using the WebQuestions dataset ($AttackNum = 5$).}
\label{fig14}
\end{figure}

\noindent {\bf{SDBA is even more powerful in transformer-based models:}} Fig.~\ref{fig11} illustrates the results of backdoor attacks on the GPT-2 model without any defenses, demonstrating that our SDBA attack exhibits significantly higher backdoor durability compared to other attacks. Subsequently, Fig.~\ref{fig12} presents the outcomes of backdoor attacks against two defense mechanisms in the GPT-2 model, showing a substantial improvement in BA over Neurotoxin, which we consider a significant achievement relative to existing backdoor research.\\
\indent Similar to Table~\ref{table2}, Table~\ref{table3} displays the $Lifespan$ results for various $\tau$ ratios, proving that SDBA poses a substantial threat as a backdoor even in transformer-based models like GPT-2. We investigated the $Lifespan$ in the GPT-2 model, as we did with the LSTM model, and found that SDBA consistently recorded longer $Lifespan$ than Neurotoxin across all experiments. Notably, the red-highlighted sections clearly show that backdoors in GPT-2 persist much longer than in LSTM, indicating that SDBA is particularly effective in transformer-based models and surpasses the performance of existing backdoor attacks.\\

\noindent {\bf{SDBA does not damage the benign model:}}  
Table~\ref{table4} shows the MA results for the model without attack, the Baseline, Neurotoxin, and SDBA. The results indicate that our SDBA does not degrade the performance of the benign model, suggesting that the backdoor attack remains stealthy and is not easily detected.\\

\noindent {\bf{Effectiveness of SDBA across Diverse NLP Tasks:}} 
Fig.~\ref{fig13} illustrates the impact of backdoor attacks on Sentiment Analysis datasets using the LSTM model without any defense mechanisms. LSTM was chosen for this evaluation due to its suitability for the dataset sizes involved, while larger models were not utilized. While the Sentiment140 dataset showed limited vulnerability to the attacks, the IMDB dataset demonstrated the superior effectiveness of our SDBA attack. Specifically, for the IMDB dataset, SDBA achieved strong backdoor durability compared to other attack methods.\\
\indent To evaluate the task generalizability of SDBA, we apply the same critical layer analysis from Section~\ref{section4b} to the T5 model. T5 is a representative transformer-based model with an encoder-decoder architecture, specialized in NLP tasks such as question answering and machine translation, making it a suitable candidate for evaluating the task generalizability of SDBA. The analysis revealed that the embedding layer, the MLP layers in the encoder and decoder, and the Self-Attention layers in the decoder act as critical layers for backdoor attacks. In particular, we found that applying 10\% masking only to the decoder MLP layers yields the most effective attack configuration.\\ 
\indent Based on this analysis, we conducted a Question Answering experiment using the T5 model to assess the performance of SDBA. Fig.~\ref{fig14} shows the impact of SDBA under this setting without any defense mechanisms. Experimental results on the WebQuestions dataset indicate that SDBA achieved higher BA and longer $Lifespan$ than Neurotoxin, demonstrating that SDBA can be effectively generalized across different task types and model architectures.\\

\begin{figure}[!t]
\centering
\includegraphics[width=3in]{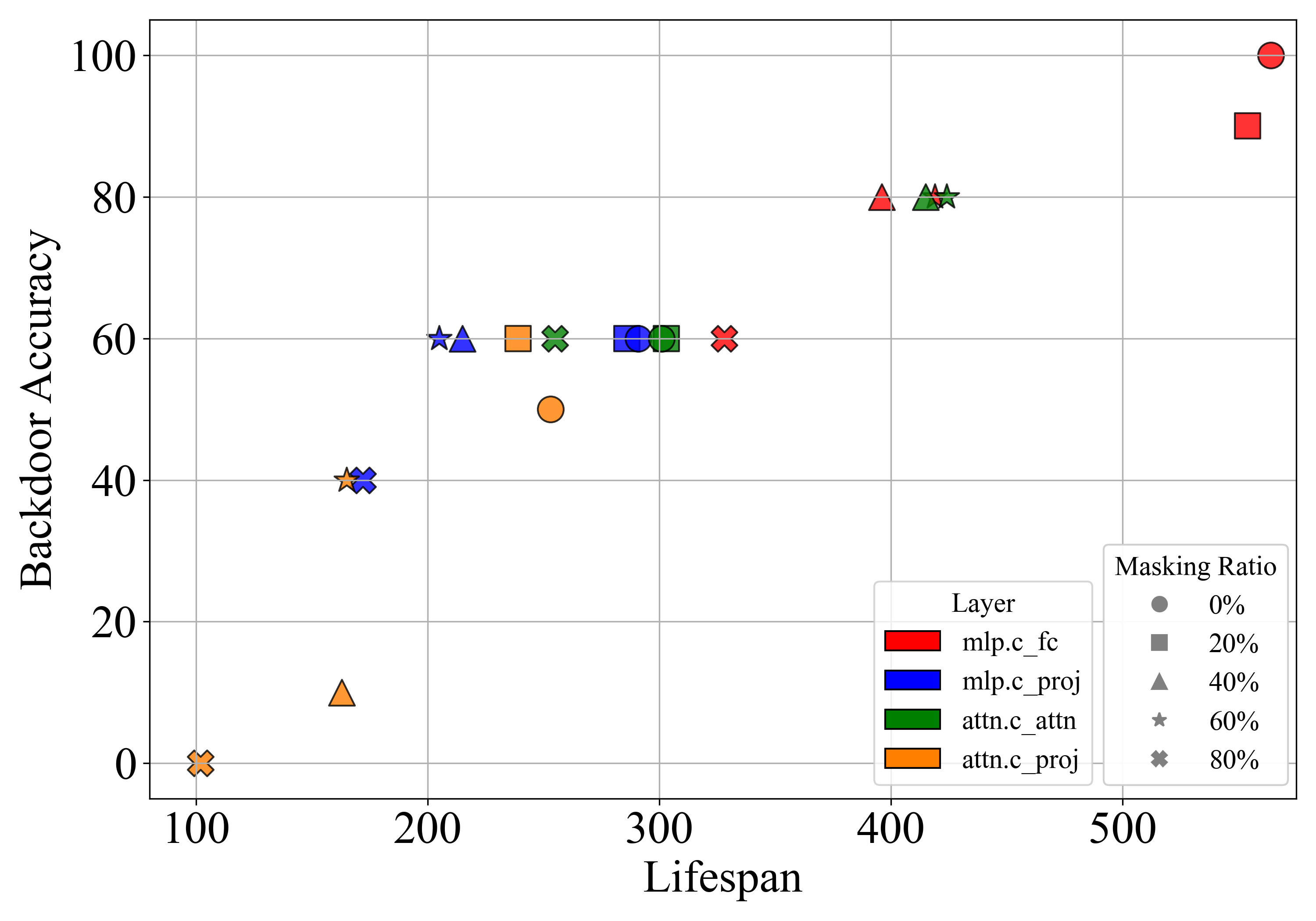}
\caption{Ablation study evaluating the impact of injection configuration on the performance of SDBA in the GPT-2 model.}
\label{fig15}
\end{figure}

\noindent {\bf{Ablation Study on Attack Scenarios:}} 
To quantitatively analyze the key attack configuration factors that influence the effectiveness of SDBA, we conduct an ablation study in a GPT-2 model environment using the Reddit dataset. Specifically, we vary the target layer for backdoor injection and the gradient masking ratio within each layer to evaluate how these parameters affect the performance of SDBA. The evaluation metrics include (1) the $Lifespan$ of the backdoor after 30 attack rounds ($\tau=50$), and (2) the BA measured at round 200 following the injection. We compare the relative performance of each experimental configuration and visually present the results for both metrics. The corresponding results are illustrated in Fig.~\ref{fig15}.\\
\indent As a result, our experimental findings show that injecting the backdoor into the $mlp.c\_fc$ layer of the MLP module with a gradient masking ratio of 1.0 yields the best performance in terms of both $Lifespan$ and BA. In contrast, the configuration targeting the $attn.c\_proj$ layer exhibits the lowest performance across both metrics. These results indicate that the effectiveness of SDBA is highly sensitive to the choice of target layer and masking intensity, thereby highlighting the critical role of component-level design in determining the success of the attack.\\

\section{Discussion} \label{section6}
To create the most powerful backdoor attack for NLP models, we conduct a detailed analysis of key factors influencing backdoor effectiveness. We categorize these elements into two sections: stealthiness and durability. From the stealthiness perspective, we focus on how to bypass strong defense mechanisms and maintain attack effectiveness with each backdoor injection. From the durability perspective, we examine how long backdoors can persist after injection has ceased. \\
\indent To address these issues, we investigate the primary vulnerabilities of each target model. For the LSTM model, backdoors were injected into the $ih$ and $hh$ layers using a layer-wise gradient masking mechanism. We empirically found that the $ih$ layer is the most critical for backdoor injection, but injecting backdoors into the $ih$ layer alone is insufficient to ensure durability. Due to the recurrent nature of LSTM, we also inject backdoors into the $hh$ layer, which stores contextual information. Additionally, top-$k\%$ gradient masking is applied to enhance durability, showing that less masking improved the $ih$ layer’s performance, while the $hh$ layer performs best without masking. The results indicate that selecting appropriate target layers plays a crucial role in achieving optimal backdoor performance.\\
\indent In the GPT-2 model, backdoors are injected into the $mlp.c\_fc$ layer, which functions as a key component for capturing input patterns and possesses high representational capacity. Unlike in the LSTM model, full poisoning of only the $mlp.c\_fc$ layer (without top-$k\%$ gradient masking) is the most durable approach for GPT-2. This method reduces the likelihood of detection and ensures that the backdoor persists longer. Furthermore, we apply the same analysis to the T5 model to evaluate how our attack is effective for other tasks such as Question Answering. The results show that our attack is more effective than existing attacks.\\
\indent We also examine the additional detectability of the proposed method. While SDBA applies layer-wise gradient masking and top-$k$\% gradient masking to zero out certain gradients, this pattern alone does not constitute a practical basis for defense. Specifically, this paper assumes a scenario where the attacker intervenes after the FL training has already converged. In such a case, benign clients may also generate sparse or low-magnitude gradients due to factors such as low learning rates or non-i.i.d. local data. Consequently, detection strategies based solely on identifying sparse gradients are prone to degrading the performance of FL, undermining their reliability in practice. However, we consider the possibility of such detection, because our SDBA sets the local gradient values to exactly zero. To evade this potential detection risk while retaining the performance of backdoors, we mask the gradient using non-zero values sampled from  $\mathcal{N}(0, \sigma_l)$, where $\mathcal{N}$ is a normal distribution and $\sigma_l$ denotes the standard deviation for the layer $l$ for each module containing the given coordinate. \\
\indent To estimate exact statistical values, the attacker builds a surrogate model trained on their held-out benign dataset in $\mathcal{P}_m$. This surrogate model is not synchronized with the global model in FL, but it is necessary to generate statistically plausible gradients that can both evade detection and preserve the performance of SDBA. \\
\indent As a result, although such defense approaches are theoretically feasible, SDBA can evade them by leveraging the surrogate model with additional computational costs. However, to the best of our knowledge, there exists no practical defense mechanism capable of detecting a single malicious client solely by comparing the global model with each local update in real-world FL scenarios. We conclude that this highlights a limitation in the generalizability of such detection strategies to various types of attacks, and also underscores the scalability challenges and computational costs associated with scanning all coordinates of local models in every training round. Therefore, in our primary attack scenario presented in Section~\ref{section3}, we adopt zero-gradient masking as the default configuration for SDBA to reduce computational overhead while maintaining effectiveness.

\section{Conclusion} \label{section7}
We proposed a novel backdoor attack method called SDBA, a stealthy and durable backdoor attack specifically designed for NLP tasks. Through a comprehensive layer-wise analysis, SDBA identifies the layers most vulnerable to backdoor injection, fulfilling both stealth and durability requirements. SDBA maintains stealthiness and is capable of evading even six representative defense mechanisms. Furthermore, by applying top-$k$\% gradient masking within specific layers, we significantly enhance the durability of the backdoor. Our experiments demonstrate that SDBA not only outperforms existing backdoor methods but also achieves greater durability than previously established attacks considered to be highly resilient.\\
\indent Future work should focus on developing robust countermeasures to combat such stealthy and durable backdoor attacks. Furthermore, from a defense perspective, our findings reveal significant threats to FL systems.\\
\indent Additionally, as mentioned in this paper, distance-based metrics employed in defense mechanisms such as Multi-Krum and FLAME are generally not applied to transformer-based models with a larger number of parameters than LSTM, due to the efficiency constraints of these methods. In summary, there is a growing need to counter increasingly stealthy and durable attacks, alongside the challenge of developing efficient defenses suitable for transformer-based models with significantly more parameters than RNN-based models such as LSTM. This highlights the need for more powerful and efficient defenses, not only for NLP tasks but also for transformer-based models.

\bibliographystyle{IEEEtran}
\bibliography{IEEEabrv, references}

\begin{thebibliography}{10}
\providecommand{\url}[1]{#1}
\csname url@samestyle\endcsname
\providecommand{\newblock}{\relax}
\providecommand{\bibinfo}[2]{#2}
\providecommand{\BIBentrySTDinterwordspacing}{\spaceskip=0pt\relax}
\providecommand{\BIBentryALTinterwordstretchfactor}{4}
\providecommand{\BIBentryALTinterwordspacing}{\spaceskip=\fontdimen2\font plus
\BIBentryALTinterwordstretchfactor\fontdimen3\font minus \fontdimen4\font\relax}
\providecommand{\BIBforeignlanguage}[2]{{%
\expandafter\ifx\csname l@#1\endcsname\relax
\typeout{** WARNING: IEEEtran.bst: No hyphenation pattern has been}%
\typeout{** loaded for the language `#1'. Using the pattern for}%
\typeout{** the default language instead.}%
\else
\language=\csname l@#1\endcsname
\fi
#2}}
\providecommand{\BIBdecl}{\relax}
\BIBdecl

\bibitem{fedavg}
B.~McMahan, E.~Moore, D.~Ramage, S.~Hampson, and B.~A. y~Arcas, ``Communication-efficient learning of deep networks from decentralized data,'' in \emph{Artificial intelligence and statistics}.\hskip 1em plus 0.5em minus 0.4em\relax PMLR, 2017, pp. 1273--1282.

\bibitem{gdpr}
P.~Voigt and A.~Von~dem Bussche, ``The eu general data protection regulation (gdpr),'' \emph{A Practical Guide, 1st Ed., Cham: Springer International Publishing}, vol.~10, no. 3152676, pp. 10--5555, 2017.

\bibitem{ccpa}
E.~L. Harding, J.~J. Vanto, R.~Clark, L.~Hannah~Ji, and S.~C. Ainsworth, ``Understanding the scope and impact of the california consumer privacy act of 2018,'' \emph{Journal of Data Protection \& Privacy}, vol.~2, no.~3, pp. 234--253, 2019.

\bibitem{gboard}
T.~Yang, G.~Andrew, H.~Eichner, H.~Sun, W.~Li, N.~Kong, D.~Ramage, and F.~Beaufays, ``Applied federated learning: Improving google keyboard query suggestions,'' \emph{arXiv preprint arXiv:1812.02903}, 2018.

\bibitem{healthcare}
J.~Li, Y.~Meng, L.~Ma, S.~Du, H.~Zhu, Q.~Pei, and X.~Shen, ``A federated learning based privacy-preserving smart healthcare system,'' \emph{IEEE Transactions on Industrial Informatics}, vol.~18, no.~3, 2021.

\bibitem{autonomous_vehicle}
Y.~Li, X.~Tao, X.~Zhang, J.~Liu, and J.~Xu, ``Privacy-preserved federated learning for autonomous driving,'' \emph{IEEE Transactions on Intelligent Transportation Systems}, vol.~23, no.~7, pp. 8423--8434, 2021.

\bibitem{howtobackdoor}
E.~Bagdasaryan, A.~Veit, Y.~Hua, D.~Estrin, and V.~Shmatikov, ``How to backdoor federated learning,'' in \emph{International conference on artificial intelligence and statistics}.\hskip 1em plus 0.5em minus 0.4em\relax PMLR, 2020, pp. 2938--2948.

\bibitem{chameleon}
Y.~Dai and S.~Li, ``Chameleon: Adapting to peer images for planting durable backdoors in federated learning,'' in \emph{International Conference on Machine Learning}.\hskip 1em plus 0.5em minus 0.4em\relax PMLR, 2023, pp. 6712--6725.

\bibitem{image_backdoor_attack1}
X.~Lyu, Y.~Han, W.~Wang, J.~Liu, B.~Wang, K.~Chen, Y.~Li, J.~Liu, and X.~Zhang, ``Coba: Collusive backdoor attacks with optimized trigger to federated learning,'' \emph{IEEE Transactions on Dependable and Secure Computing}, 2024.

\bibitem{image_backdoor_attack2}
C.~Shi, S.~Ji, X.~Pan, X.~Zhang, M.~Zhang, M.~Yang, J.~Zhou, J.~Yin, and T.~Wang, ``Towards practical backdoor attacks on federated learning systems,'' \emph{IEEE Transactions on Dependable and Secure Computing}, 2024.

\bibitem{image_backdoor_attack3}
K.~Wei, J.~Li, M.~Ding, C.~Ma, Y.-S. Jeon, and H.~V. Poor, ``Covert model poisoning against federated learning: Algorithm design and optimization,'' \emph{IEEE Transactions on Dependable and Secure Computing}, 2023.

\bibitem{chatbot}
E.~Adamopoulou and L.~Moussiades, ``Chatbots: History, technology, and applications,'' \emph{Machine Learning with applications}, vol.~2, p. 100006, 2020.

\bibitem{machine_translation}
J.~Zhang, C.~Zong \emph{et~al.}, ``Deep neural networks in machine translation: An overview.'' \emph{IEEE Intell. Syst.}, vol.~30, no.~5, pp. 16--25, 2015.

\bibitem{sentiment_analysis}
W.~Medhat, A.~Hassan, and H.~Korashy, ``Sentiment analysis algorithms and applications: A survey,'' \emph{Ain Shams engineering journal}, vol.~5, no.~4, pp. 1093--1113, 2014.

\bibitem{neurotoxin}
Z.~Zhang, A.~Panda, L.~Song, Y.~Yang, M.~Mahoney, P.~Mittal, R.~Kannan, and J.~Gonzalez, ``Neurotoxin: Durable backdoors in federated learning,'' in \emph{International Conference on Machine Learning}.\hskip 1em plus 0.5em minus 0.4em\relax PMLR, 2022, pp. 26\,429--26\,446.

\bibitem{lstm}
S.~Hochreiter, ``Long short-term memory,'' \emph{Neural Computation MIT-Press}, 1997.

\bibitem{gpt2}
A.~Radford, J.~Wu, R.~Child, D.~Luan, D.~Amodei, I.~Sutskever \emph{et~al.}, ``Language models are unsupervised multitask learners,'' \emph{OpenAI blog}, vol.~1, no.~8, p.~9, 2019.

\bibitem{t5}
\BIBentryALTinterwordspacing
C.~Raffel, N.~Shazeer, A.~Roberts, K.~Lee, S.~Narang, M.~Matena, Y.~Zhou, W.~Li, and P.~J. Liu, ``Exploring the limits of transfer learning with a unified text-to-text transformer,'' \emph{Journal of Machine Learning Research}, vol.~21, no. 140, pp. 1--67, 2020. [Online]. Available: \url{http://jmlr.org/papers/v21/20-074.html}
\BIBentrySTDinterwordspacing

\bibitem{multi_krum}
P.~Blanchard, E.~M. El~Mhamdi, R.~Guerraoui, and J.~Stainer, ``Machine learning with adversaries: Byzantine tolerant gradient descent,'' \emph{Advances in neural information processing systems}, vol.~30, 2017.

\bibitem{normclip_weakdp_pgd}
Z.~Sun, P.~Kairouz, A.~T. Suresh, and H.~B. McMahan, ``Can you really backdoor federated learning?'' \emph{arXiv preprint arXiv:1911.07963}, 2019.

\bibitem{flame}
T.~D. Nguyen, P.~Rieger, R.~De~Viti, H.~Chen, B.~B. Brandenburg, H.~Yalame, H.~M{\"o}llering, H.~Fereidooni, S.~Marchal, M.~Miettinen \emph{et~al.}, ``$\{$FLAME$\}$: Taming backdoors in federated learning,'' in \emph{31st USENIX Security Symposium (USENIX Security 22)}, 2022, pp. 1415--1432.

\bibitem{fedprox}
T.~Li, A.~K. Sahu, M.~Zaheer, M.~Sanjabi, A.~Talwalkar, and V.~Smith, ``Federated optimization in heterogeneous networks,'' \emph{Proceedings of Machine learning and systems}, vol.~2, pp. 429--450, 2020.

\bibitem{aggregation1}
Y.~Cai, W.~Ding, Y.~Xiao, Z.~Yan, X.~Liu, and Z.~Wan, ``Secfed: A secure and efficient federated learning based on multi-key homomorphic encryption,'' \emph{IEEE Transactions on Dependable and Secure Computing}, 2023.

\bibitem{aggregation2}
R.~Xu, B.~Li, C.~Li, J.~Joshi, S.~Ma, T.~Zhou, J.~Dong, and J.~Li, ``Tapfed: Threshold secure aggregation for privacy-preserving federated learning,'' \emph{IEEE Transactions on Dependable and Secure Computing}, 2024.

\bibitem{aggregation3}
Y.~Wang, A.~Zhang, S.~Wu, and S.~Yu, ``Vosa: Verifiable and oblivious secure aggregation for privacy-preserving federated learning,'' \emph{IEEE Transactions on Dependable and Secure Computing}, vol.~20, no.~5, pp. 3601--3616, 2022.

\bibitem{aggregation4}
X.~Tang, M.~Shen, Q.~Li, L.~Zhu, T.~Xue, and Q.~Qu, ``Pile: Robust privacy-preserving federated learning via verifiable perturbations,'' \emph{IEEE Transactions on Dependable and Secure Computing}, vol.~20, no.~6, pp. 5005--5023, 2023.

\bibitem{badnet}
T.~Gu, K.~Liu, B.~Dolan-Gavitt, and S.~Garg, ``Badnets: Evaluating backdooring attacks on deep neural networks,'' \emph{IEEE Access}, vol.~7, pp. 47\,230--47\,244, 2019.

\bibitem{difficult_detect}
G.~Baruch, M.~Baruch, and Y.~Goldberg, ``A little is enough: Circumventing defenses for distributed learning,'' \emph{Advances in Neural Information Processing Systems}, vol.~32, 2019.

\bibitem{clean_label_backdoor}
\BIBentryALTinterwordspacing
L.~Gan, J.~Li, T.~Zhang, X.~Li, Y.~Meng, F.~Wu, Y.~Yang, S.~Guo, and C.~Fan, ``Triggerless backdoor attack for {NLP} tasks with clean labels,'' in \emph{Proceedings of the 2022 Conference of the North American Chapter of the Association for Computational Linguistics: Human Language Technologies}, M.~Carpuat, M.-C. de~Marneffe, and I.~V. Meza~Ruiz, Eds.\hskip 1em plus 0.5em minus 0.4em\relax Seattle, United States: Association for Computational Linguistics, Jul. 2022, pp. 2942--2952. [Online]. Available: \url{https://aclanthology.org/2022.naacl-main.214/}
\BIBentrySTDinterwordspacing

\bibitem{hidden_trigger}
\BIBentryALTinterwordspacing
X.~Pan, M.~Zhang, B.~Sheng, J.~Zhu, and M.~Yang, ``Hidden trigger backdoor attack on {NLP} models via linguistic style manipulation,'' in \emph{31st USENIX Security Symposium (USENIX Security 22)}.\hskip 1em plus 0.5em minus 0.4em\relax Boston, MA: USENIX Association, Aug. 2022, pp. 3611--3628. [Online]. Available: \url{https://www.usenix.org/conference/usenixsecurity22/presentation/pan-hidden}
\BIBentrySTDinterwordspacing

\bibitem{edgecase}
H.~Wang, K.~Sreenivasan, S.~Rajput, H.~Vishwakarma, S.~Agarwal, J.-y. Sohn, K.~Lee, and D.~Papailiopoulos, ``Attack of the tails: Yes, you really can backdoor federated learning,'' \emph{Advances in Neural Information Processing Systems}, vol.~33, pp. 16\,070--16\,084, 2020.

\bibitem{byzantine}
L.~Lamport, R.~Shostak, and M.~Pease, ``The byzantine generals problem,'' in \emph{Concurrency: the works of leslie lamport}, 2019, pp. 203--226.

\bibitem{dp_sgd}
M.~Abadi, A.~Chu, I.~Goodfellow, H.~B. McMahan, I.~Mironov, K.~Talwar, and L.~Zhang, ``Deep learning with differential privacy,'' in \emph{Proceedings of the 2016 ACM SIGSAC conference on computer and communications security}, 2016, pp. 308--318.

\bibitem{critical_layer}
H.~Zhuang, M.~Yu, H.~Wang, Y.~Hua, J.~Li, and X.~Yuan, ``Backdoor federated learning by poisoning backdoor-critical layers,'' \emph{arXiv preprint arXiv:2308.04466}, 2023.

\bibitem{webquestions}
\BIBentryALTinterwordspacing
J.~Berant, A.~Chou, R.~Frostig, and P.~Liang, ``Semantic parsing on {F}reebase from question-answer pairs,'' in \emph{Proceedings of the 2013 Conference on Empirical Methods in Natural Language Processing}.\hskip 1em plus 0.5em minus 0.4em\relax Seattle, Washington, USA: Association for Computational Linguistics, Oct. 2013, pp. 1533--1544. [Online]. Available: \url{https://www.aclweb.org/anthology/D13-1160}
\BIBentrySTDinterwordspacing

\bibitem{geva2021}
\BIBentryALTinterwordspacing
M.~Geva, R.~Schuster, J.~Berant, and O.~Levy, ``Transformer feed-forward layers are key-value memories,'' in \emph{Proceedings of the 2021 Conference on Empirical Methods in Natural Language Processing}, M.-F. Moens, X.~Huang, L.~Specia, and S.~W.-t. Yih, Eds.\hskip 1em plus 0.5em minus 0.4em\relax Online and Punta Cana, Dominican Republic: Association for Computational Linguistics, Nov. 2021, pp. 5484--5495. [Online]. Available: \url{https://aclanthology.org/2021.emnlp-main.446/}
\BIBentrySTDinterwordspacing

\bibitem{rome}
K.~Meng, D.~Bau, A.~Andonian, and Y.~Belinkov, ``Locating and editing factual associations in gpt,'' \emph{Advances in neural information processing systems}, vol.~35, pp. 17\,359--17\,372, 2022.

\bibitem{reddit}
H.~B. McMahan, D.~Ramage, K.~Talwar, and L.~Zhang, ``Learning differentially private recurrent language models,'' \emph{arXiv preprint arXiv:1710.06963}, 2017.

\bibitem{sentiment140}
A.~Go, R.~Bhayani, and L.~Huang, ``Twitter sentiment classification using distant supervision,'' \emph{CS224N project report, Stanford}, vol.~1, no.~12, p. 2009, 2009.

\bibitem{imdb}
A.~Maas, R.~E. Daly, P.~T. Pham, D.~Huang, A.~Y. Ng, and C.~Potts, ``Learning word vectors for sentiment analysis,'' in \emph{Proceedings of the 49th annual meeting of the association for computational linguistics: Human language technologies}, 2011, pp. 142--150.

\end{thebibliography}

\begin{IEEEbiography}[{\includegraphics[width=1in,height=1.25in,clip,keepaspectratio]{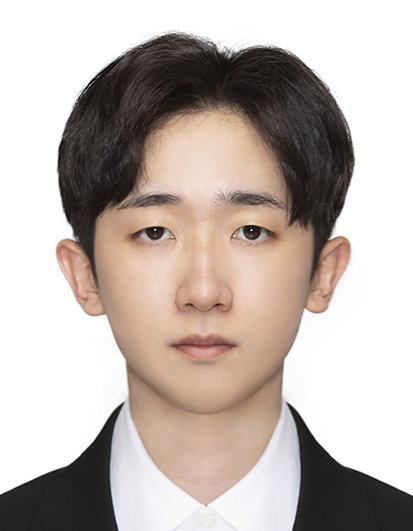}}]{Minyeong Choe}
received the B.S. degree from the Department of Information and Communication Engineering, Chosun University, Gwangju, South Korea, in 2025. He is currently an M.S. student in the same department at Chosun University. His research interests include private and secure federated learning, data privacy, and AI security.
\end{IEEEbiography}

\begin{IEEEbiography}[{\includegraphics[width=1in,height=1.25in,clip,keepaspectratio]{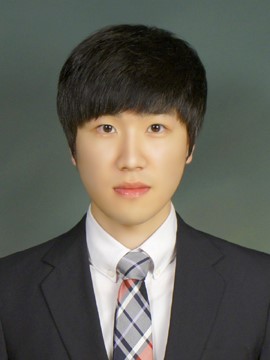}}]{Cheolhee Park}
received the B.S. degree from the Department of Applied Mathematics, Kongju National University, Gongju, South Korea, in 2014, and the M.S. and Ph.D. degrees from the Department of Mathematics, Kongju National University in 2017 and 2021, respectively. He joined Electronics and Telecommunications Research Institute, Daejeon, South Korea, in 2021, where he is currently working as a Researcher. His research interests include data privacy, differential privacy, machine learning, deep learning, AI security, and network security.
\end{IEEEbiography}

\begin{IEEEbiography}[{\includegraphics[width=1in,height=1.25in,clip,keepaspectratio]{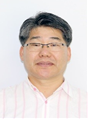}}]{Changho Seo}
 received the B.S., M.S., and Ph.D. degrees from the Department of Mathematics, Korea University, South Korea, in 1990, 1992, and 1996, respectively. He is currently a Professor with the Department of Convergence Science, Kongju National University, South Korea. His research interests include cryptography, information security, data privacy protection, and AI security.
\end{IEEEbiography}

\begin{IEEEbiography}[{\includegraphics[width=1in,height=1.25in,clip,keepaspectratio]{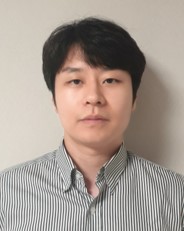}}]{Hyunil Kim}
 received the B.S. degree from the Department of Applied Mathematics, Kongju National University, Gongju, South Korea, in 2014, and the M.S. and Ph.D. degrees from the Department of Convergence Science, Kongju National University in 2016 and 2019, respectively. He was a Postdoctoral Researcher with DGIST, Daegu, South Korea, from 2020 to 2022. He is currently a Assistant Professor with the Department of Artificial Intelligence and Software Engineering, Chosun University, South Korea. His research interests include private and secure federated learning, data privacy, and AI security.
\end{IEEEbiography}

\end{document}